\documentclass[10pt,journal]{IEEEtran}

\usepackage[numbers,sort&compress]{natbib}
\usepackage{times}
\usepackage{epsfig} 
\usepackage{epstopdf}
\usepackage{graphicx}
\usepackage{amsmath}
\usepackage{amssymb}
\usepackage{subfigure}
\usepackage[boxed]{algorithm2e}
\usepackage{url}
\usepackage{color}
\usepackage{amsmath,bm}
\usepackage{threeparttable}
\usepackage{booktabs}
\usepackage{multirow}

\makeatletter

\newcommand{\Rmnum}[1]{\expandafter\@slowromancap\romannumeral #1@}
\makeatother

\graphicspath{{figure/}}

\begin{document}
\title{Latent Fingerprint Registration via Matching Densely Sampled Points}

\author{Shan~Gu,~\IEEEmembership{Student Member,~IEEE,}
Jianjiang~Feng,~\IEEEmembership{Member,~IEEE,}
Jiwen~Lu,~\IEEEmembership{Senior~Member,~IEEE,}
and~Jie~Zhou,~\IEEEmembership{Senior~Member,~IEEE} 
\IEEEcompsocitemizethanks{\IEEEcompsocthanksitem

This work was supported in part by the National Natural Science Foundation of China under Grants 61976121, 61622207 and 61527808, and in part by the Shenzhen Fundamental Research Fund (subject arrangement) under Grant JCYJ20170412170438636. (Corresponding author:
Jianjiang Feng.)

Shan Gu, Jiwen Lu, and Jie Zhou are with Department of Automation, Beijing National Research Center for Information Science and Technology, Tsinghua University, Beijing 100084, China (e-mail: gus16@mails.tsinghua.edu.cn; lujiwen@tsinghua.edu.cn; jzhou@tsinghua.edu.cn).

Jianjiang Feng is with Department of Automation, Beijing National Research Center for Information Science and Technology, Tsinghua University, Beijing 100084, China, and also with the Graduate School at Shenzhen, Tsinghua University, Shenzhen 518055, China (e-mail: jfeng@tsinghua.edu.cn).
}
}
\markboth{IEEE TRANSACTIONS ON Information Forensics and
Security,~Vol.~X, No.~X,~XXXXX~XXXX}{Shell \MakeLowercase{\textit{et
al.}}: Bare Demo of IEEEtran.cls for Journals}

\maketitle
\thispagestyle{empty}

\begin{abstract}

Latent fingerprint matching is a very important but unsolved problem. As a key step of fingerprint matching, fingerprint registration has a great impact on the recognition performance.
Existing latent fingerprint registration approaches are mainly based on establishing correspondences between minutiae, and hence will certainly fail when there are no sufficient number of extracted minutiae due to small fingerprint area or poor image quality. Minutiae extraction has become the bottleneck of latent fingerprint registration.
In this paper,  we propose a non-minutia latent fingerprint registration method  which estimates the spatial  transformation between a pair of fingerprints through a dense fingerprint patch alignment and matching procedure.  
Given a pair of fingerprints to match, we bypass the minutiae extraction step and take uniformly sampled points as key points. Then the proposed patch alignment and matching algorithm compares all pairs of sampling points and produces their similarities along with alignment parameters. Finally, a set of consistent correspondences are found by spectral clustering.
Extensive experiments on NIST27  database and MOLF database show that the proposed method achieves the state-of-the-art registration performance, especially under challenging conditions. Code is made publicly available at: \emph{https://github.com/Gus233/Latent-Fingerprint-Registration}.

\end{abstract}


\section{Introduction}

Latent fingerprints have been playing a vital role in identifying suspects \cite{maltoni2009handbook}. Up to now, manual feature extraction and matching is still indispensable in latent fingerprint matching. Recently, automatic latent fingerprint feature extraction and matching has become a research focus in the field of fingerprint recognition, aiming to reduce the workload of fingerprint experts.
Despite of many published work on this topic, the recognition performance degrades greatly when the latent fingerprint has small effective area, low image quality or large distortion \cite{sankaran2014latent}. Therefore, there is still large room for improvement in latent fingerprint matching.

\begin{figure}[t]
\begin{center}	
	\subfigure[manually marked minutiae]{\includegraphics[width=0.49\linewidth]{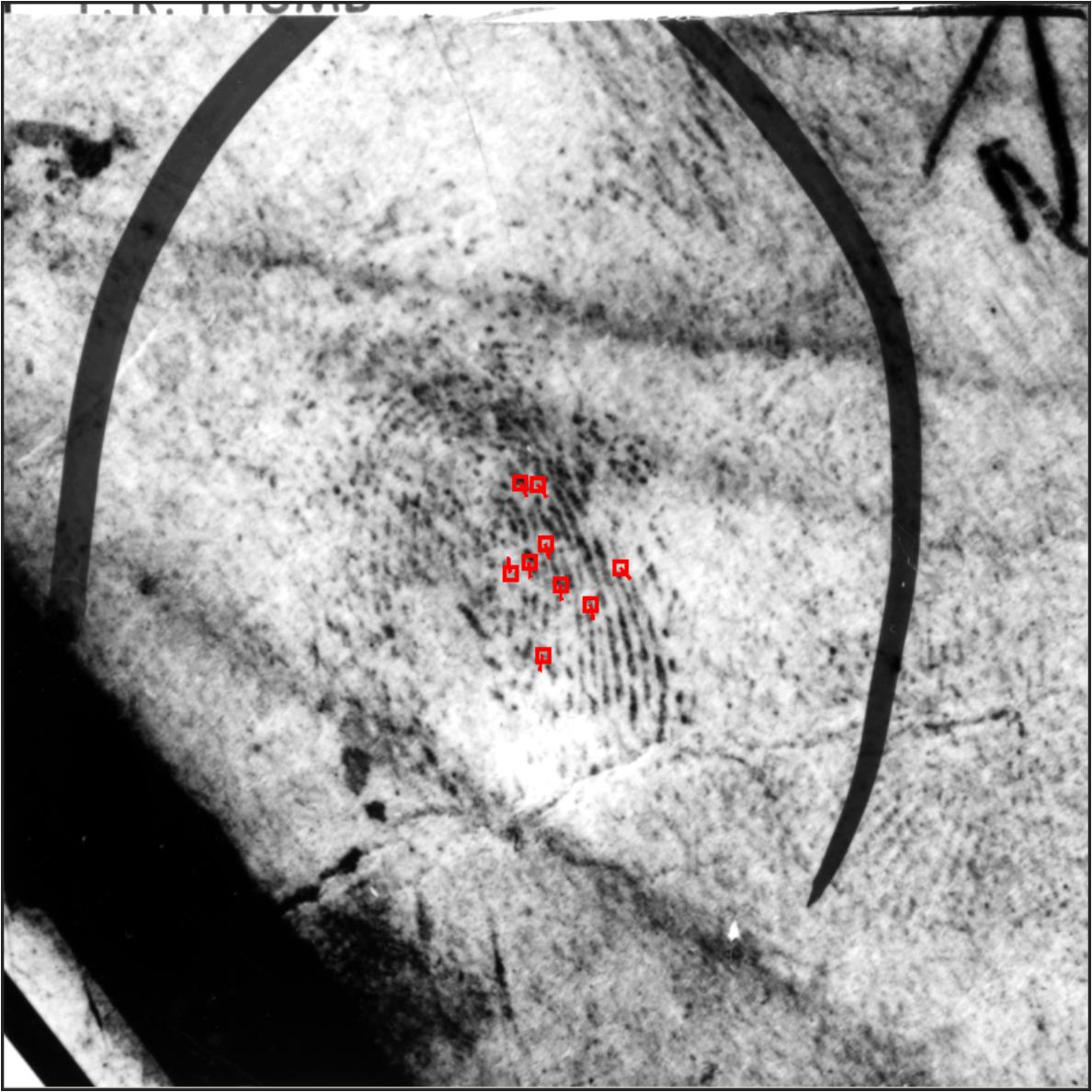}}
	\subfigure[minutiae estimated by FingerNet]{\includegraphics[width=0.49\linewidth]{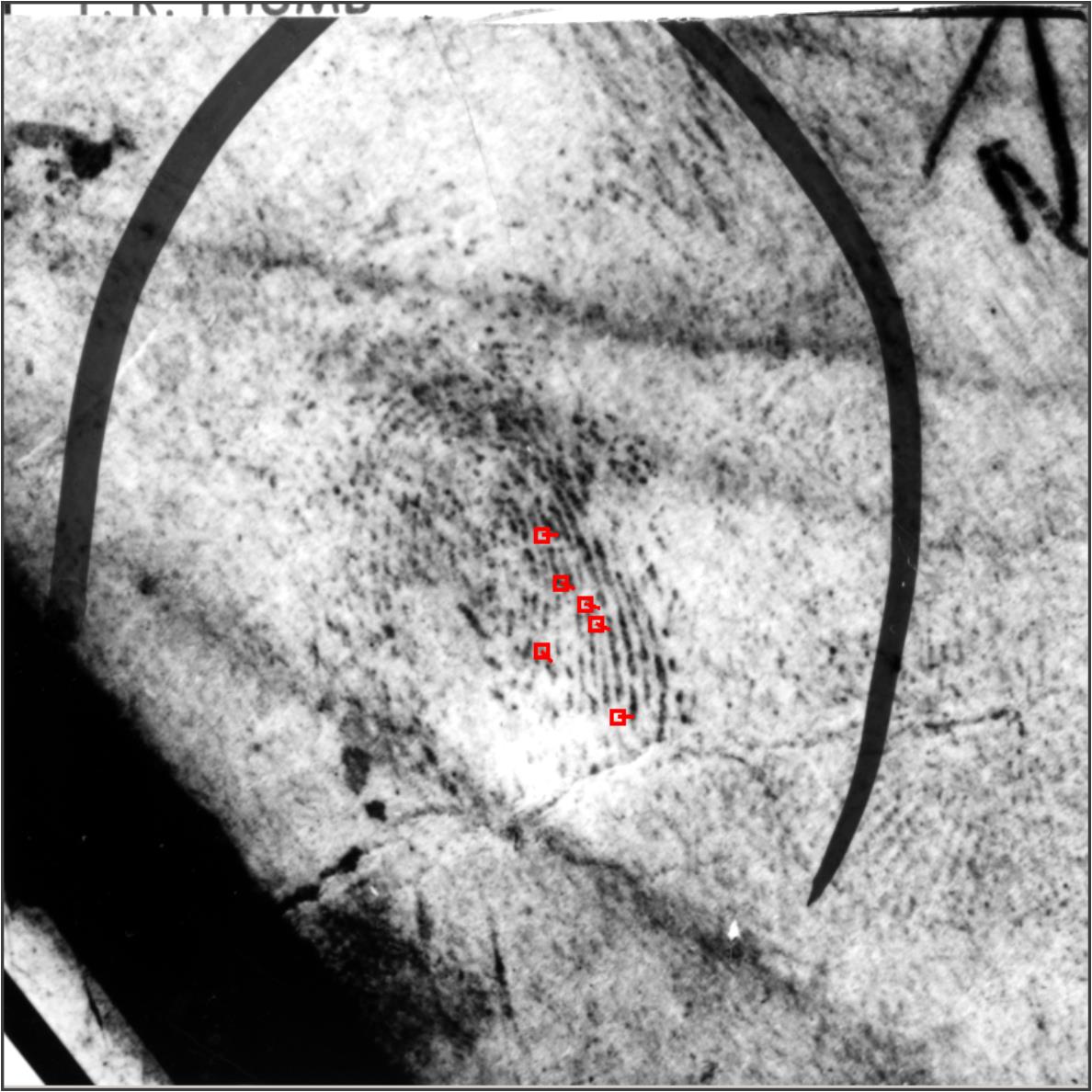}}
	\subfigure[sampling point correspondences obtained by the proposed registration approach]{\includegraphics[width=0.99\linewidth]{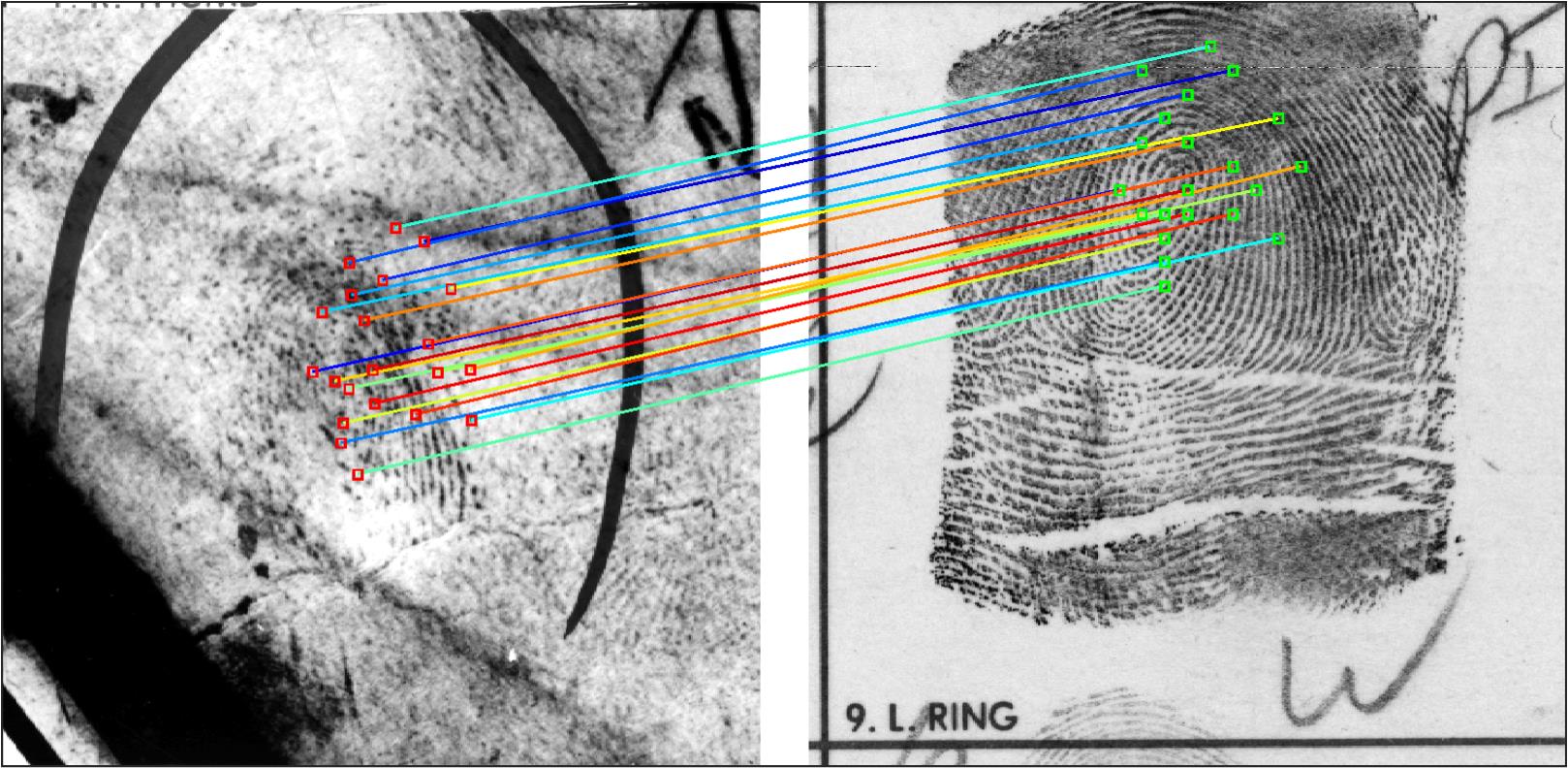}}
\end{center}
	\caption{A challenging case where the latent fingerprint and the mated rolled fingerprint can be successfully registered by the proposed approach. 
	 The image quality in the latent fingerprint is  so poor that none of minutiae extracted by FingerNet \cite{tang2017fingernet} is correct.  The proposed method can find sampling point correspondences  under such challenging situation.
 }
\label{fig: example}
\end{figure}

\begin{figure*}
\begin{center}
	\includegraphics[width=\linewidth]{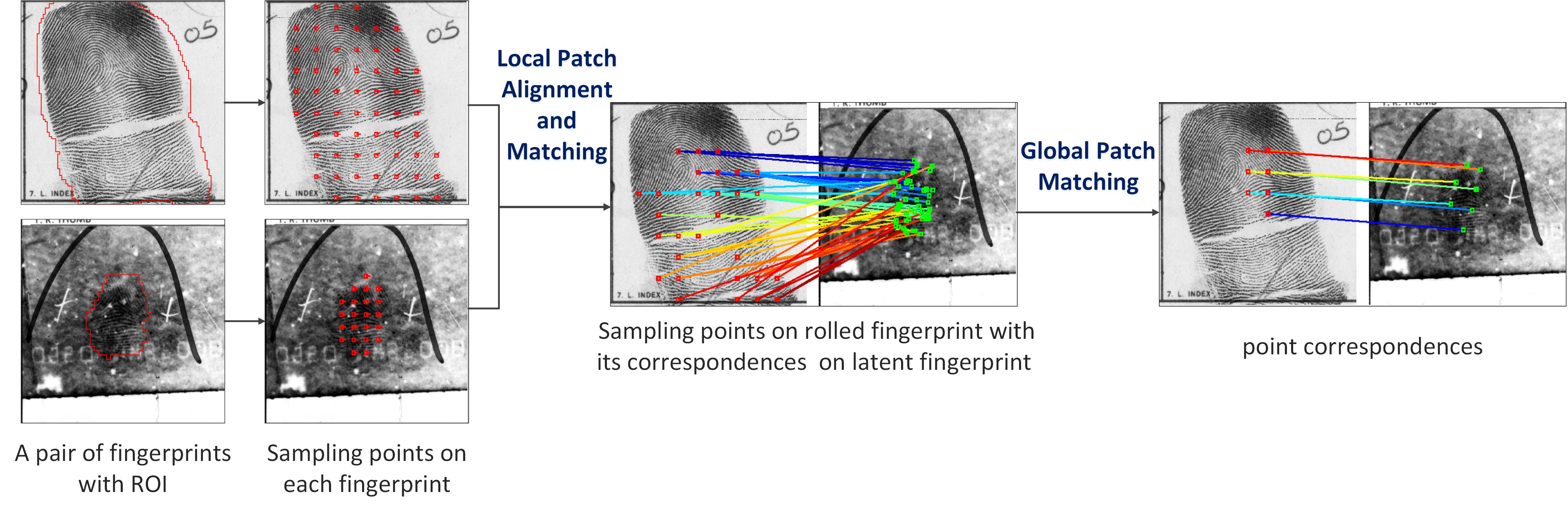}
\end{center}
	\caption{The flowchart of the proposed fingerprint registration algorithm. %
	The coarse stage of coarse-to-fine registration scheme is shown here.  Sampling points are first sampled uniformly in the ROI on each fingerprint.    Then local patch alignment and matching module compares all pairs of sampling points, aligns local patches finely,  and finds potential correspondences. Global patch matching module is finally applied to find a set of compatible corresponding sampling points among all potential correspondences.
  }
\label{fig: systemflowchart}
\end{figure*}

In  latent fingerprint matching algorithms, accurate fingerprint registration is of vital importance for the recognition performance. Minutiae based methods \cite{bazen2003fingerprint,ross2005deformable} are  most commonly used to establish correspondences across images, which first extract minutiae from each fingerprint, then extract minutiae descriptors, and finally use them to find reliable matches. Among the three steps,  accurate extraction of minutiae is the basis, which will greatly affect the accuracy of the subsequent steps.
However,  the requirement for sufficient and accurate minutiae is usually difficult to meet for latent fingerprints with low quality.  
The estimated minutiae may be very few when the latent fingerprints have small effective area or  inaccurate in location or direction  when fingerprint images are greatly occluded by background noise. Fig. \ref{fig: example} shows an example in NIST27 database \cite{NIST27}, where none of minutiae extracted by FingerNet \cite{tang2017fingernet}, a state-of-the-art minutiae extractor,  is correct. Minutiae based registration methods will certainly fail in such cases.

In order to overcome the  limitations of minutiae based approaches, orientation field based registration \cite{krish2015pre} has been proposed considering that level 1 features are more robust to noise. However, this method greatly relies on reliable orientation  field extraction. Besides, the orientation  field is not discriminative enough for very small fingerprints.
Cao \textit{et al.} \cite{cao2018automated,cao2018end} proposed to use densely sampled points (referred to as virtual minutiae) for latent fingerprint recognition. The virtual minutiae are defined as uniformly sampled directed points  on fingerprint regions with  local ridge orientation as the direction.  With virtual minutiae, the following matching procedure is similar with minutiae based approaches.  
The virtual minutiae based method can alleviate the problems of insufficient minutiae caused by  small area or poor quality, but it still suffers from  unstable estimation of local ridge orientation, which is common in latent fingerprints. The wrong estimation of direction of virtual minutiae will greatly affect the performance of virtual minutiae descriptors since the descriptor is extracted from local image patch aligned with the direction.  
 Another limitation is that virtual minutiae are not salient features and cannot be located accurately, so the accuracy of fingerprint registration is limited by sampling intervals. To achieve registration accuracy at the pixel level, this method needs to take each pixel as a sampling point, which has very high computational complexity and is impractical.

In this paper, a new latent fingerprint registration algorithm based on dense sampling points is proposed.  In the challenging example shown in Fig. \ref{fig: example}, the proposed algorithm can still find a number of sampling point correspondences.
The proposed method can further enhance the anti-noise ability of registration algorithms and has high efficiency, which is attributed to the following designs.

The core of the proposed registration algorithm is local patch alignment and matching  (see the flowchart shown in Fig. \ref{fig: systemflowchart}).
Fingerprint is represented as dense sampling points in replace of minutiae to avoid minutiae extraction step and ensure the adequacy of key points even if fingerprints area is very small. It should be noted that  we only estimate the directions of  sampling points on rolled fingerprints. 
Then the local patch  alignment and matching algorithm  estimates the alignment parameters between image patches and computes their similarities.
Compared with   previous minutiae or virtual minutiae based registration,  we add the local patch alignment module before extracting sampling point descriptors, which is key to handle their limitations. Concretely, (1) the estimation of relative rotation between two sampling points rather than separate estimation of their own local direction helps obtain more  accurate estimated directions of sampling points on latent fingerprints. (2) The estimation of relative offset between two sampling points helps the algorithm achieve pixel-level accuracy without the need of sampling every pixel.  
In addition, the proposed local patch matching module computes the similarity of two original image patches rather than enhanced ones to ensure that the image will not be damaged by the fingerprint enhancement algorithm. In conclusion, we skip the traditional fingerprint feature extraction framework of extracting level-1 (ridge orientation and frequency) and level-2 (minutiae) features on latent fingerprints and thus avoid being affected by errors in feature extraction.

After that, potential correspondences between sampling points are obtained by comparing their similarities to a predefined threshold. Since there are still a large number of false correspondences, spectral clustering based global patch matching is employed to find a set of consistent corresponding sampling points.

Considering registration accuracy and time complexity, a  coarse-to-fine registration scheme is proposed. The registration algorithm in both stages is the same, and the precise registration takes the result of coarse registration as input. 
In coarse registration stage,  the sampling interval is large, and all pairs of sampling points on two fingerprints are considered for comparison.  In precise registration, the sampling points are denser, but each sampling point is compared only with its neighbors. 
With such a scheme, large amount of comparison in coarse registration is required but the number of sampling points is small, while in precise registration the sampling points are denser  but the number of matches required for each point decreases.

Extensive experiments on latent fingerprint registration are conducted on NIST27 database \cite{NIST27} and MOLF latent database \cite{sankaran2015multisensor}. On NIST27 database, the distances between manually marked matching minutiae after registration are used as the evaluation metric to directly evaluate the proposed registration method. Besides,  matching experiments on NIST27 and MOLF latent  fingerprint databases have been carried out to  further verify whether the proposed registration algorithm is beneficial to the matching performance.
The experimental results show that our approach performs better than existing methods, especially under challenging conditions such as heavy background noise and small fingerprint area.

The rest of the paper is organized as follows.
In Section 2, we review the related work. In section 3, we describe the proposed latent fingerprint registration method including the details of local patch alignment and matching, and  global patch matching. In Section 4, we evaluate the performance of the proposed registration method. In Section 5, we conclude the paper.

\begin{figure}[t]
\begin{center}
	\subfigure[]{\includegraphics[width=0.4\linewidth]{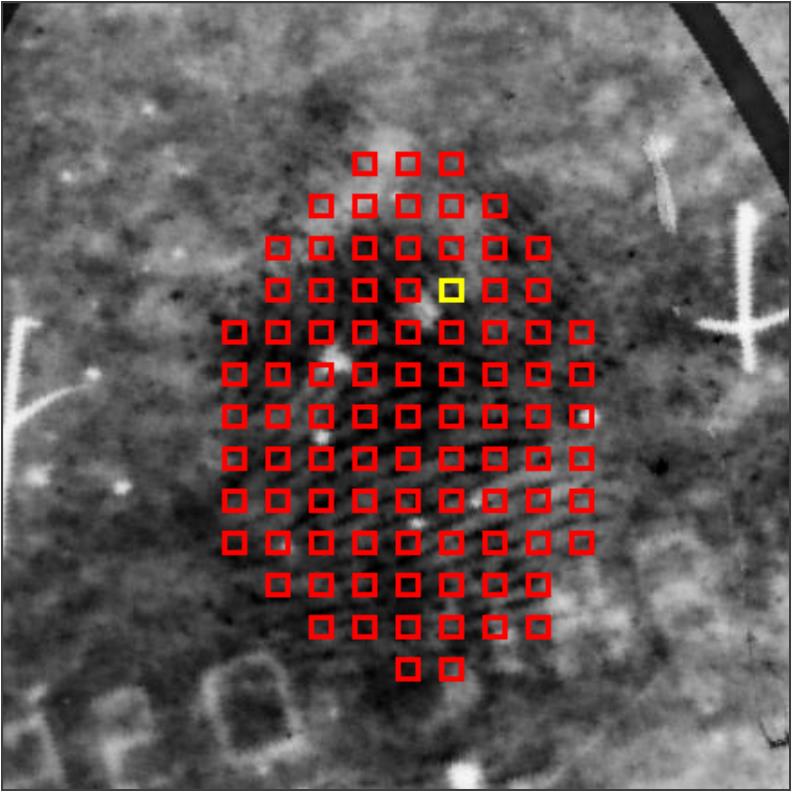}}
	\subfigure[]{\includegraphics[width=0.4\linewidth]{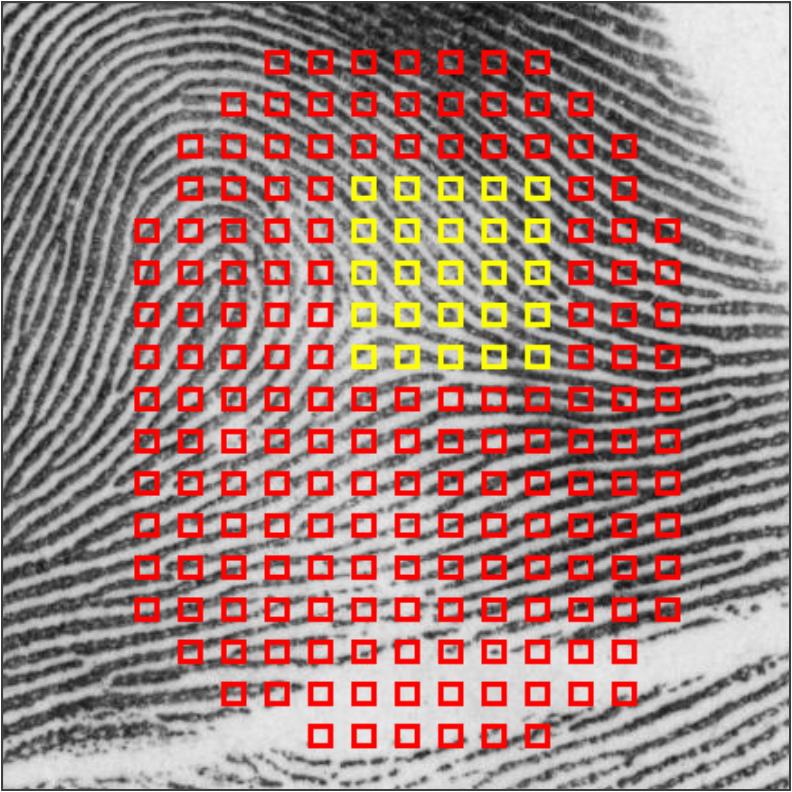}}
\end{center}
	\caption{Sampling points on a pair of fingerprints (has undergone coarse registration) in the precise registration stage. The sampling intervals on both fingerprints are $24 \times 24$. A sampling point on the latent fingerprint (marked in yellow in (a)) is only compared with neighboring sampling points (marked in yellow in (b)) on the rolled fingerprint.}
\label{fig: sampling}
\end{figure}

\section{Related Work}

All kinds of fingerprint registration algorithms can be regarded as key point-based registration algorithms when we view minutiae, ridge sampling points, image sampling points, sweat pores, and even pixels as  key points. In the following, we review several representative  work pertaining to different modules including the selection of key points, descriptors, and global registration. 

\subsection{Key Points} 
The key points in registration algorithms contain minutiae, sampling points on ridges, sampling points on image, sweat pores,  and pixels. 
Minutiae \cite{bazen2003fingerprint,ross2005deformable} are the most commonly used since it is greatly distinctive and can be extracted robustly on high quality fingerprints.  
However, minutiae are still sparse and can not generate very precise registration results when fingerprints are distorted. 

To improve the registration performance, a lot of work combines minutiae with other key points. 
In order to better deal with distortion,  sampling points on ridges \cite{ross2005fingerprint,lin2018matching}, undirected sampling points on images \cite{si2017dense}, and pixels \cite{cui20182} are aligned after initial minutiae registration.
For high resolution fingerprints, level 3 features \cite{jain2006pores} including pore and ridge contours are extracted as key points, which can boost the performance when combining with minutiae registration.
For latent fingerprints with small effective area, rare minutia features \cite{krish2019improving} are exploited to supplement the insufficient minutiae. When  latent fingerprints are  with  so low quality that minutiae and ridges may not be accurately extracted, Cao \textit{et al.} \cite{cao2018automated,cao2018end} proposed virtual minutiae (directed sampling points on images) to ensure enough key points on fingerprint area. But their direction may be mis-estimated, which would affect the matching performance.

In this paper, we adopt dense undirected sampling points on latent fingerprint images to ensure adequate key points  and do not define their absolute direction to avoid possible error of direction estimation. Different from the method in \cite{si2017dense}, our method does not require minutiae registration as initialization.

\subsection{Descriptors}
Descriptors are very important for identifying whether two key points match or not.
Minutiae descriptors have been widely researched,  and the descriptors for other key points \cite{si2017dense} are similar with them. Therefore,  we discuss minutiae descriptors in the following.

Traditional minutiae descriptors can be divided into three categories: image based, texture based, and minutiae based descriptors. 
Image based  \cite{kovacs2000fingerprint,bazen2000correlation} and texture based \cite{tico2003fingerprint,benhammadi2007fingerprint}  descriptors use image intensity or ridge orientation information, while minutiae based descriptors \cite{chen2006new,feng2008combining,xu2009fingerprint,cappelli2010minutia,iloanusi2011indexing,khodadoust2017fingerprint,li2014score} make use of the relationships between neighboring minutiae. These methods use manually designed features and thus are difficult to optimize to separate mated minutiae pairs from non-mated pairs, especially under challenging situations.

Recently, descriptors constructed by deep learning \cite{zhang2017combining,cao2018automated,song2018aggregating,cao2018end} are proposed to capture the essential characteristics of fingerprint patches. To ensure the robustness of descriptors, local image patches are aligned in advance based on their estimated  key point direction.
Experiments show their superior performance compared to traditional descriptors. However, their performance may be degraded when image patches are aligned with wrong direction. Besides, since these methods use enhanced fingerprints as input, it  may also cause performance degradation when images are enhanced with wrong ridge orientation, which is common in latent fingerprints.

In our method, we align the image patches with estimated key point directions before extracting their deep descriptors. Specially, the key point direction of latent image patch is estimated by the fusion of that of rolled image patch and the related spatial transformation. Besides, the  deep descriptors are learned from original image patches rather than enhanced ones, which avoid being affected by errors in ridge orientation estimation.

\begin{figure*}
\begin{center}
	\includegraphics[width=1\linewidth]{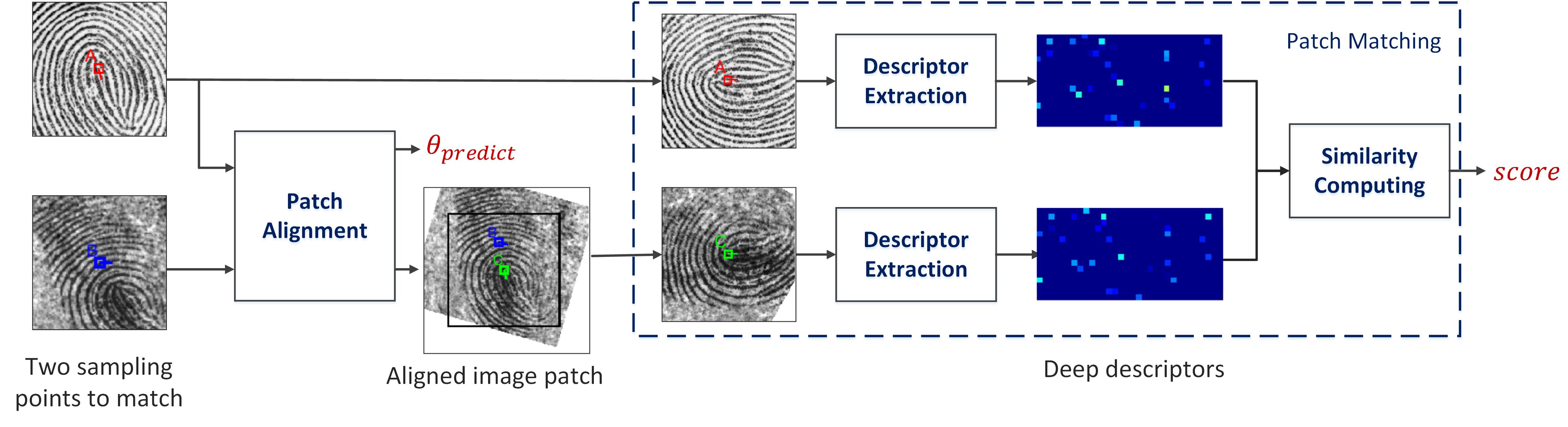}
\end{center}
	\caption{The flowchart of the proposed local patch alignment and matching algorithm. 
	Given two image patches (from latent and rolled fingerprints) centered on the sampling points $A$ and $B$, the patch alignment module predicts the spatial transformation $\theta_{predict}$ between them. The sampling point $A$ take the value of orientation map on that point  as its direction while the sampling point $B$ has no direction. 
	With the predicted transformation,  the latent fingerprint patch  are well aligned to the rolled fingerprint patch, and the adjusted sampling point $C$ on the latent fingerprint can be obtained, which is the actually possible corresponding point of $A$.
    After that, both image patches are aligned with the direction of the sampling point and represented as   deep descriptors. The similarity between descriptors is outputted to reflect how similar the two patches are.
	}
\label{fig: patchmatchingflowchart}
\end{figure*}

\subsection{Global Registration}
Minutiae based approaches usually fit a thin-plate spline model with paired minutiae as landmark points. The correspondences between minutiae are established by considering both similarity of minutiae descriptors and compatibility between minutiae pairs \cite{fu2013extended, fu2015minutia}. Cao and Jain \cite{cao2018automated} introduced the compatibility between minutiae triplets to further eliminate false correspondences. 

Many non-minutiae based methods \cite{ross2005fingerprint,si2017dense,cui20182} use minutiae registration as an initial step because minutiae are salient, and their small number leads to high matching efficiency. In the following, more correspondences are found for more precise registration. After initial registration,  Ross \textit{et al.} \cite{ross2005fingerprint}   used a thin-plate spline model to  establish correspondences between   sampling points on the ridge curves.  Si \textit{et al.} \cite{si2017dense} proposed a block matching algorithm to find dense correspondences and  minimized an energy function for global optimization. However, incorrect minutiae registration will result in incorrect initial registration of these methods, which is difficult to remedy by  the following precise registration.

Our method takes sampling points as key points and  does not use minutiae registration as initialization in order to improve the performance for low quality images and small area fingerprints. We establish the initial correspondences according to the similarities between sampling point pairs and use the pairwise geometric constraints to select a subset of consistent correspondences by spectral clustering.

\section{Proposed Registration Algorithm}

In this section, we  describe the proposed registration algorithm, including  overall framework,  local patch alignment,  local patch matching,   global patch matching,  and implementation details of these modules.

\subsection{Coarse-to-fine framework}

In order to balance registration accuracy and time complexity, the proposed approach applies a coarse-to-fine   registration framework. It contains two stages: coarse registration and precise registration. The registration process in both stages is the same, but the  setting of sampling points is different.  

In coarse registration stage, original pair of fingerprints are taken as input. The sampling intervals on rolled fingerprints are $80 \times 80$, while that on latent fingerprint  are $48 \times 48$ due to small fingerprint area.  We compare all the sampling points on latent fingerprint  with those on rolled fingerprint  to find the coarse correspondences.

In precise registration stage, we take  registered fingerprints after coarse registration as input.  The sampling intervals on both fingerprints are $24 \times 24$ considering both the accuracy and efficiency. Since the the distance between corresponding points may  not be too far after coarse registration, the sampling points on latent fingerprints are only compared with neighboring sampling points on rolled fingerprints. An example of comparison is shown in Fig. \ref{fig: sampling}.

\subsection{Local Patch Alignment}

The proposed local patch alignment and matching module aims to find the  potential correspondences. Given one pair of sampling points, this module estimates their relative spatial transformations and  computes their similarity. If the similarity  is higher than a given threshold, this pair of sampling point is considered as potential correspondence. The flowchart of the proposed local patch alignment and matching module is  illustrated in Fig. \ref{fig: patchmatchingflowchart}.   Patch matching will be described in next subsection.

Since sampling points are sampled separately on each fingerprint,  a pair of sampling points are usually not exact corresponding points even when the two patches are roughly corresponding.
Therefore, it is necessary to adjust the location and direction of sampling points on latent fingerprints to obtain the accurate correspondences. The patch alignment approach is proposed to estimate the relative spatial transformation between two image patches. Based on the estimation, the location and direction of  sampling points are adjusted for better alignment. 

\begin{figure}
\begin{center}
	\includegraphics[width=\linewidth]{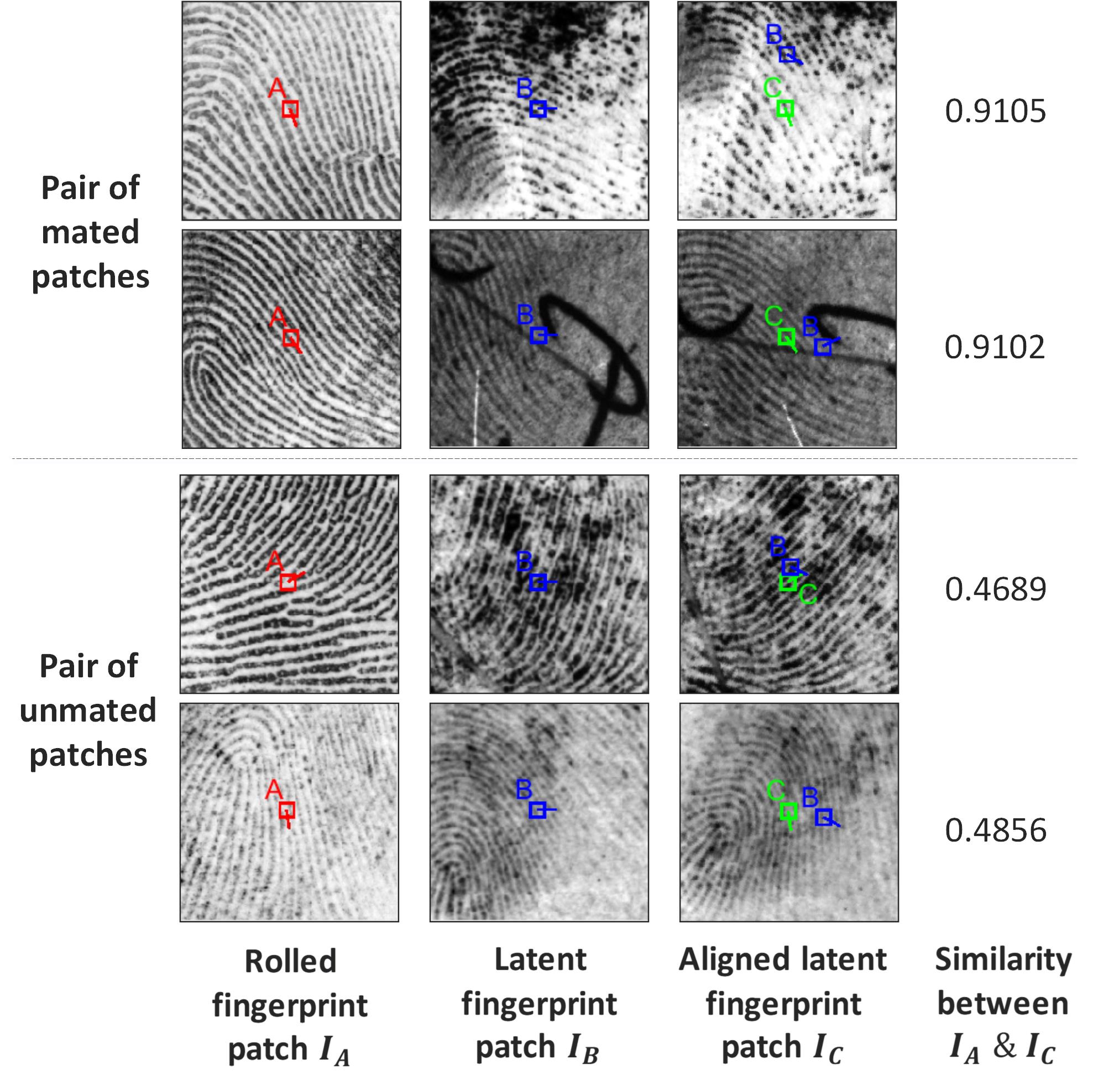}
\end{center}
	\caption{Results of the proposed patching alignment and matching algorithm on mated and unmated fingerprint patches.}
\label{fig: patchpairs}
\end{figure}

To illustrate the effect of patch alignment, Fig. \ref{fig: patchpairs} shows  results of the proposed patch alignment and matching algorithm  on mated and unmated  pairs of fingerprint patches.  For mated patches $I_A$ and $I_B$ from a pair of rolled fingerprint and latent fingerprint, they  still have a large difference in translation due to large sampling interval.   They  also have   relative rotation  due to the rotation of fingerprints.  If  their  similarity is computed without better alignment, it will  be very low and misclassified as non-mated.  But with the patch alignment module, the similarity between  $I_A$ and the aligned patch $I_C$ can reflect more accurate matching results.
On the other hand, if $I_A$ and $I_B$ are not mated patches, no matter what the estimation of transformation parameters is,  the descriptors extracted from $I_A$ and the aligned patch $I_C$ are dissimilar.

We train a Siamese network to estimate the geometric transformation following \cite{rocco2017convolutional}.  The network architecture contains three parts. The feature extraction network extract  feature maps from image patches centered on two sampling points. Taking these two feature maps as input, a matching layer matches them  and outputs a correlation map. With the correlation map, the translation and rotation parameters are estimated with a regression network.

\subsubsection{Model Architecture}
\begin{figure}[t]
\begin{center}
		\includegraphics[width=\linewidth]{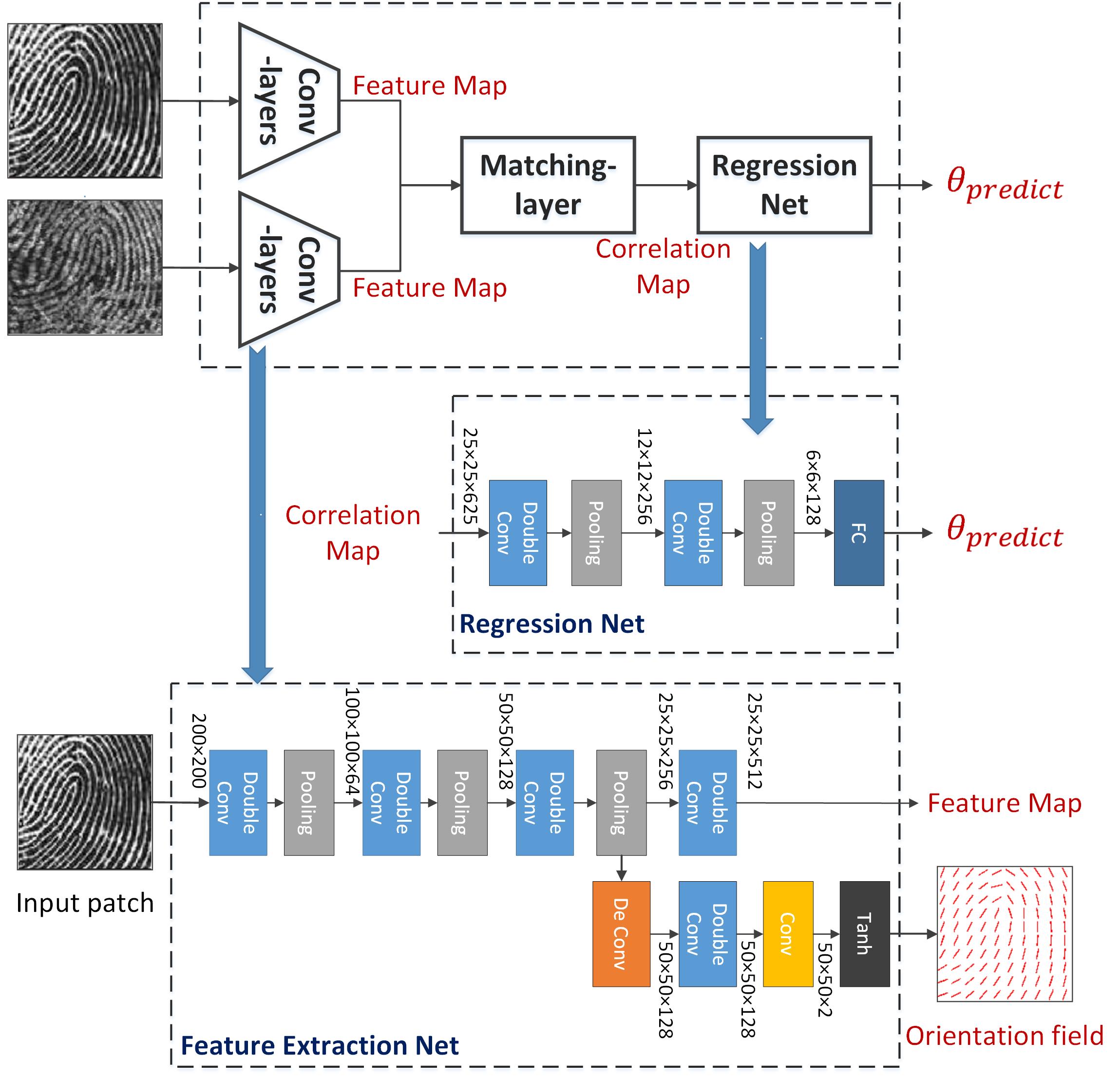}
\end{center}
	\caption{The model architecture of the proposed patch alignment algorithm.}
\label{fig: matchingmodel}
\end{figure}

The flowchart of patch alignment algorithm is illustrated in Fig. \ref{fig: matchingmodel}. For the feature extraction network, we use the standard VGG-13 network and crop it before the pool4 layer. It has four double-conv blocks,  and the first three are followed by a max pooling layer. Each double-conv block contains two conv layers where each conv layer followed by a BatchNorm layer and a ReLU layer. 
To overcome different appearances of input images, we add a ridge orientation supervision to help the network pay more attention to fingerprint ridges rather than the image background. We add the supervision layers at pool3, and let the network learns to estimate the orientation field. The supervision layers are made of a deconv layer, a double-conv block, a conv layer, and a Tanh layer.

After each image patch produce its feature map, the matching network is developed to combine two feature maps to a single one for the subsequent parameter estimation. We use a matching procedure similar to \cite{rocco2017convolutional}. The correlation layer computes all pairs of similarities between feature maps and is followed by similarity normalization. 
Given the correlation map, a regression map is applied to estimate the translation and rotation parameters between two image patches. The regression network is composed of two double-conv blocks and one fully connected layer. The output is 3-D vector, which indicates the relative translation in horizontal and vertical directions, and the relative rotation angle, respectively.

\subsubsection{Loss Function} We train the patch alignment network  in a fully supervised manner. For a pair of input image patches, the ground truth transformation parameter is given by $\theta_{gt}=[dx, dy,da]$, indicates the relative translation and rotation angle. With the estimated parameter $\theta_{pred}$, we use MSE loss to evaluate the parameter regression error:
\begin{equation}
\begin{split}
L_{para}(\theta_{gt}, \theta_{pred}) = \frac{1}{N} \sum_{i=1}^{N}\left\|\theta_{gt}^i-\theta_{pred}^i\right\|^2
\end{split}
\end{equation}

For the ridge orientation estimation, we also use the MSE loss for convenience. For an image patch of size $H \times W$,  the corresponding ground truth orientation map $M_{gt}$  is a two-channel image map concatenating the cosine map  $cos(2O)$ and sine map $sin(2O)$ of origin orientation field $O$ whose size is $\frac{H}{8} \times \frac{W}{8}$.
The MSE loss for one input image patch is
\begin{equation}
\begin{split}
L_{ori}(M_{gt}, M_{pred}) = \frac{1}{N} \sum_{i=1}^{N}\left\|M_{gt}^i-M_{pred}^i\right\|^2
\end{split}
\end{equation}

The final loss is a weighted sum of the loss of transformation estimation and the total loss of  ridge orientation estimation for two input image patches, which is
\begin{equation}
\begin{split}
L_{match} = L_{para}+ \lambda_{match} (L_{ori}^1+L_{ori}^2)
\end{split}
\end{equation}
where $\lambda_{match}$ is the weighted coefficient.

\begin{figure}[t]
\begin{center}
		\includegraphics[width=\linewidth]{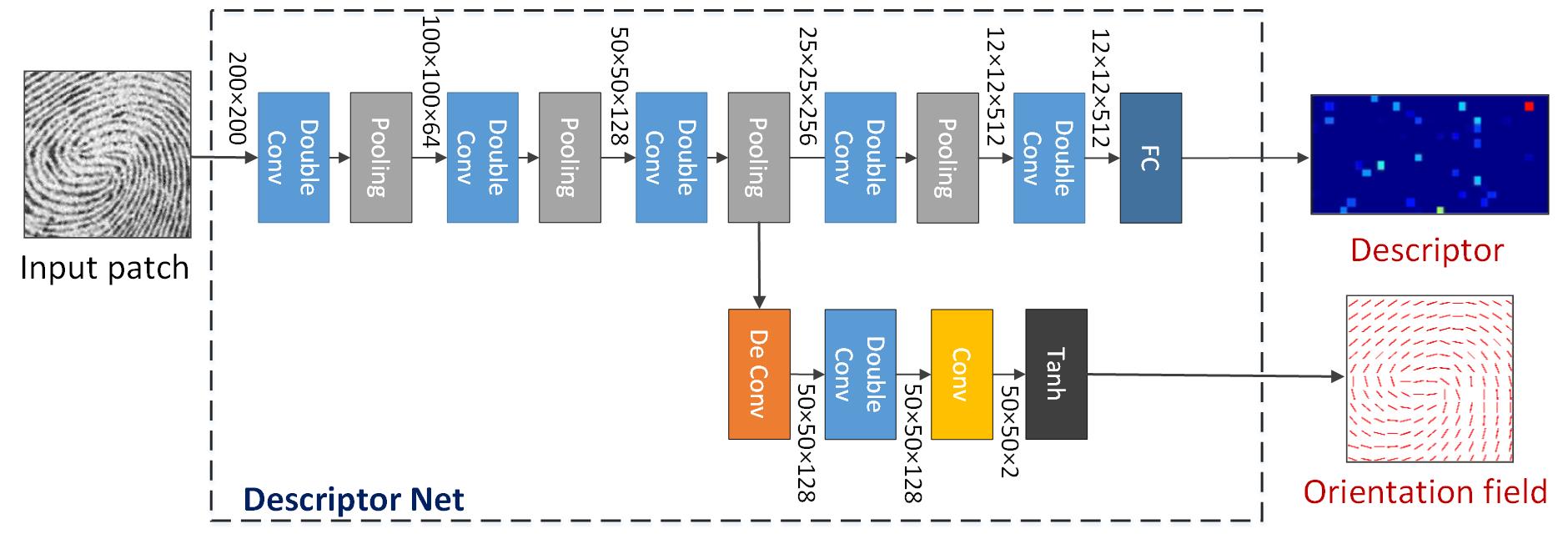}
\end{center}
	\caption{The model architecture of the proposed deep descriptor.}
\label{fig: descriptormodel}
\end{figure}

\subsubsection{Simulation of Training Data}
Training models in the fully supervised manner requires  pairs of image patches and their corresponding transformation parameters. We prepare the training data by applying synthetic transformations to real pairs of latent and rolled fingerprints, which makes it easier to get large amount of data for the CNN training. See Section D for information on the used fingerprint database.

To synthesize the training data, the latent fingerprints are required to register to their paired rolled ones. 
We register two fingerprints based on the MCC descriptor \cite{cappelli2010minutia} and spectral clustering \cite{leordeanu2005spectral}. 
The MCC is used to measure the similarity between all possible minutiae pairs, and spectral clustering is used to obtain the mated minutiae pairs. Based on the paired minutiae, Thin Plate Spline (TPS) \cite{bookstein1989principal} model is applied to approximate the distortion field. 
In order to ensure the correctness of training samples, we apply quality control manually to choose only examples whose registration is correct.
Then we cut pairs of image patches whose size is $200\times 200$ randomly on the registered image pairs. Only pairs of image patches with common foreground area greater than a predefined threshold are preserved. In this way, we can obtain multiple pairs of aligned image patches from each pair of latent and rolled fingerprints. 
For a pair of image patch $I_A$ and $I_B$, random translation and rotation are applied to them, and the transformed patches are named as $\hat{I_A}$ and $\hat{I_B}$. 
We take the patch pair $(I_A,\hat{I_B})$ and $(I_A,\hat{I_A})$ as training pairs. The image patches in the latter pair are both from rolled fingerprints, which can make the network easier to train.

Specifically, we generate two training datasets for the coarse and precise registration, respectively. The parameter dimensions of the transformations are  different in two datasets. In coarse registration, the range of random translation is $[-50,50]$ pixels,  and that of rotation is  $[-40,40]$ degrees, while in  precise registration the range of random translation is set as $[-25,25]$ pixels and that of rotation is set as $[-20,20]$ degrees. When training models,  the parameters of translation are transformed to $[-1,1]$.

 \begin{figure*}
\begin{center}
	\includegraphics[width=\linewidth]{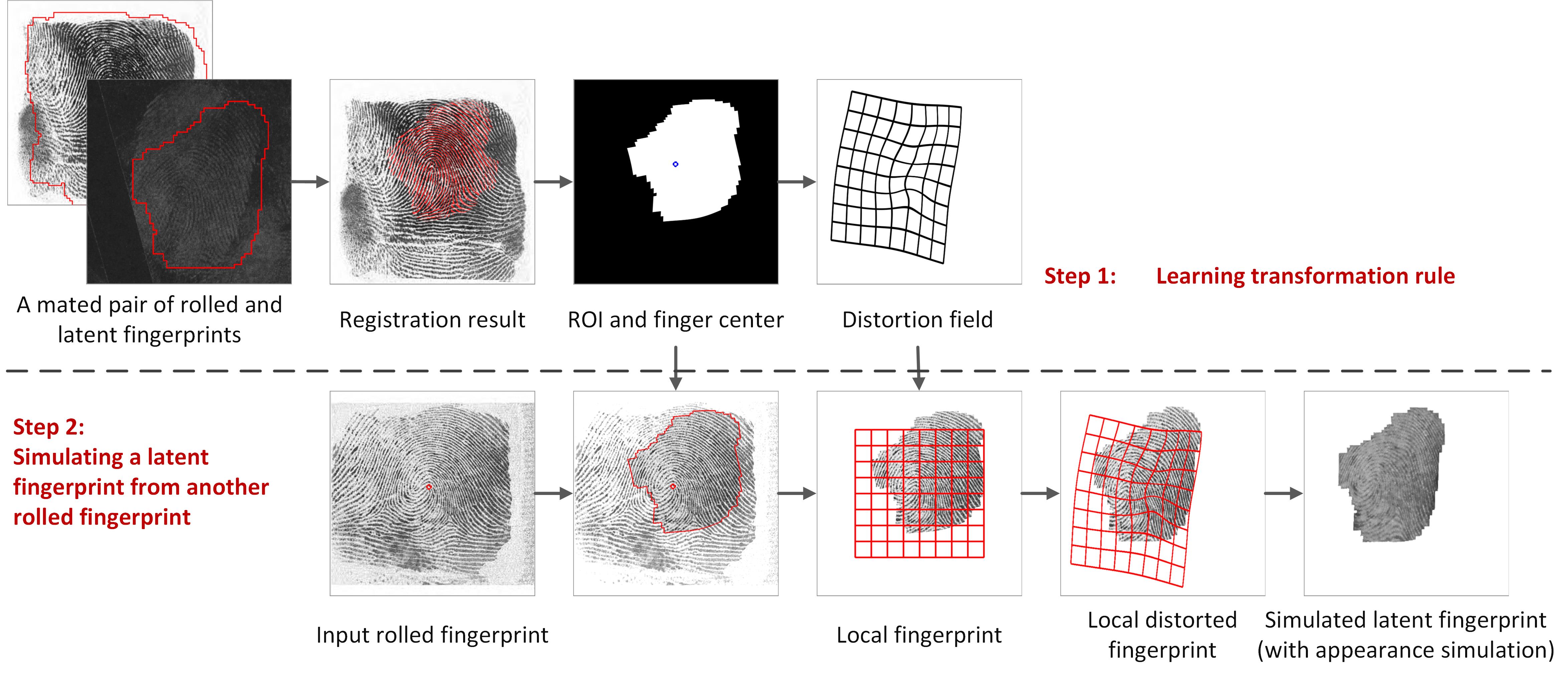}
\end{center}
	\caption{The proposed fingerprint simulation algorithm consists of two steps. }
\label{fig: simulation}
\end{figure*}

\subsection{Local Patch Matching}

The proposed local patch matching procedure aims to compute the similarity between a pair of aligned image patches. 
To compute the similarity, we extract deep descriptor from image patches  centered on each sampling point with the point direction as the horizontal positive direction, and take the distance between two descriptors as the patch similarity.  

We train a Siamese network to extract deep descriptors. Although minutiae descriptors based on depth learning  \cite{cao2018automated,zhang2017combining,cao2018end,song2018aggregating} have been proposed, our main contribution is to propose a dedicated process to address the problem of lacking of ground truth data.    
Simulating multiple fingerprints from rolled fingerprints can  simultaneously solve the problem about the lack of samples and correspondence information.

\subsubsection{Model Architecture}

The VGGNet-13 architecture is adopted for the feature embedding. Similar with the feature extraction network in the proposed patch alignment algorithm, the supervision is added after pool3 to let the ridge orientation as supervisory information. After all the conv layers, the FC-BN structure is applied to obtain a 512-dimensional feature, which is used as the descriptor. The concrete architecture is illustrated in Fig. \ref{fig: descriptormodel}.

\subsubsection{Loss Function}
Similar with the loss in local patch alignment algorithm, the loss function here contains two parts, the contrastive loss $L_{contrastive}$ which minimizes the distance between positive pairs while maximizes that of negative pairs, and  the MSE loss $L_{ori}$ which evaluates the estimation accuracy of ridge orientation.
The total loss is  
\begin{equation}
\begin{split}
L_{simi} = L_{contrastive}+ \lambda_{simi} (L_{ori}^1+L_{ori}^2)
\end{split}
\end{equation}
where $\lambda_{simi}$ is the weight coefficient.

\subsubsection{Simulation of Training Data}
To train the descriptor learning network, multiple samples are required for the same key point to enhance the generalization of the network, and the correspondence between key points should be known. 
However, existing public latent fingerprint databases usually have only two fingerprints for one finger. That is to say, one key point at most have  two patches.  In addition, to obtain the key point correspondences, additional key point matching is required, which would make the data preparation more complex and cannot guarantee the accuracy and completeness of  correspondences.

To obtain large latent fingerprint database, an efficient alternative is to simulate fingerprint images \cite{during2019anisotropic,cao2018fingerprint}. We therefore propose a fingerprint simulation method  to efficiently obtain more reliable training data. Using this method, multiple image patches from the same key point and the correspondence between them can be obtained simultaneously. 
The basic idea is that, the transformation rule is analyzed in advance and then transferred to other fingerprints to simulate more images. 
In the following, we first introduce the proposed fingerprint simulation method and then the generation of key point patches.

\textbf{Fingerprint Simulation}
The simulation process includes two steps: learning transformation rule and simulating fingerprints from rolled fingerprints with the transformation rule. One example of the detailed steps is shown in Fig. \ref{fig: simulation}.

In the first step, we register latent fingerprints with corresponding rolled fingerprints to obtain the region of interest (ROI) and distortion field. 
The ROI is used to simulate small valid area of fingerprints while the distortion field is used to simulate the deformation.

We register two fingerprints in a similar manner with that in the data simulation of the proposed local patch alignment algorithm. The registration is conducted by fitting the TPS model with matched minutiae, which are found by spectral clustering method with similarities of minutiae pairs computed by MCC descriptor.
After registering latent fingerprints to rolled fingerprints, the intersection of the ROI of registered latent fingerprints and that of rolled fingerprints is seen as the common ROI. To better locate the relative position of the common ROI, the fingerprint center of the rolled fingerprint is also estimated. It is defined as the upper core of a fingerprint, which can be estimated by the VeriFinger \cite{NeuroTech}.

It should be noted that we use the MCC-based minutiae matching algorithm to  efficiently get pairs of training samples. Although the algorithm is not perfect, our experimental results show that the performance of the deep minutiae descriptor trained based on these training samples is significantly better than that of MCC.

In the second step, given one rolled fingerprint, a corresponding fingerprint can be simulated with the given distortion field and ROI.  The ROI is firstly applied to the rolled fingerprint tocut a local fingerprint area. To ensure the right absolute position, the center recorded with ROI should be in advance aligned with the center of the rolled fingerprint. Then we apply the distortion field on the clipped local fingerprints. 

Finally, we apply the CycleGAN \cite{zhu2017unpaired} to change the fingerprint appearance to simulate the gray-scale ridge pattern in various  fingerprints. Multiple pairs of registered latent and rolled fingerprints are applied to train the model. 
In this way,  the incompleteness, distortion, and different appearance of fingerprints are all simulated.

\begin{figure*}[t]
\begin{center}
	\subfigure[]{\includegraphics[width=0.45\linewidth]{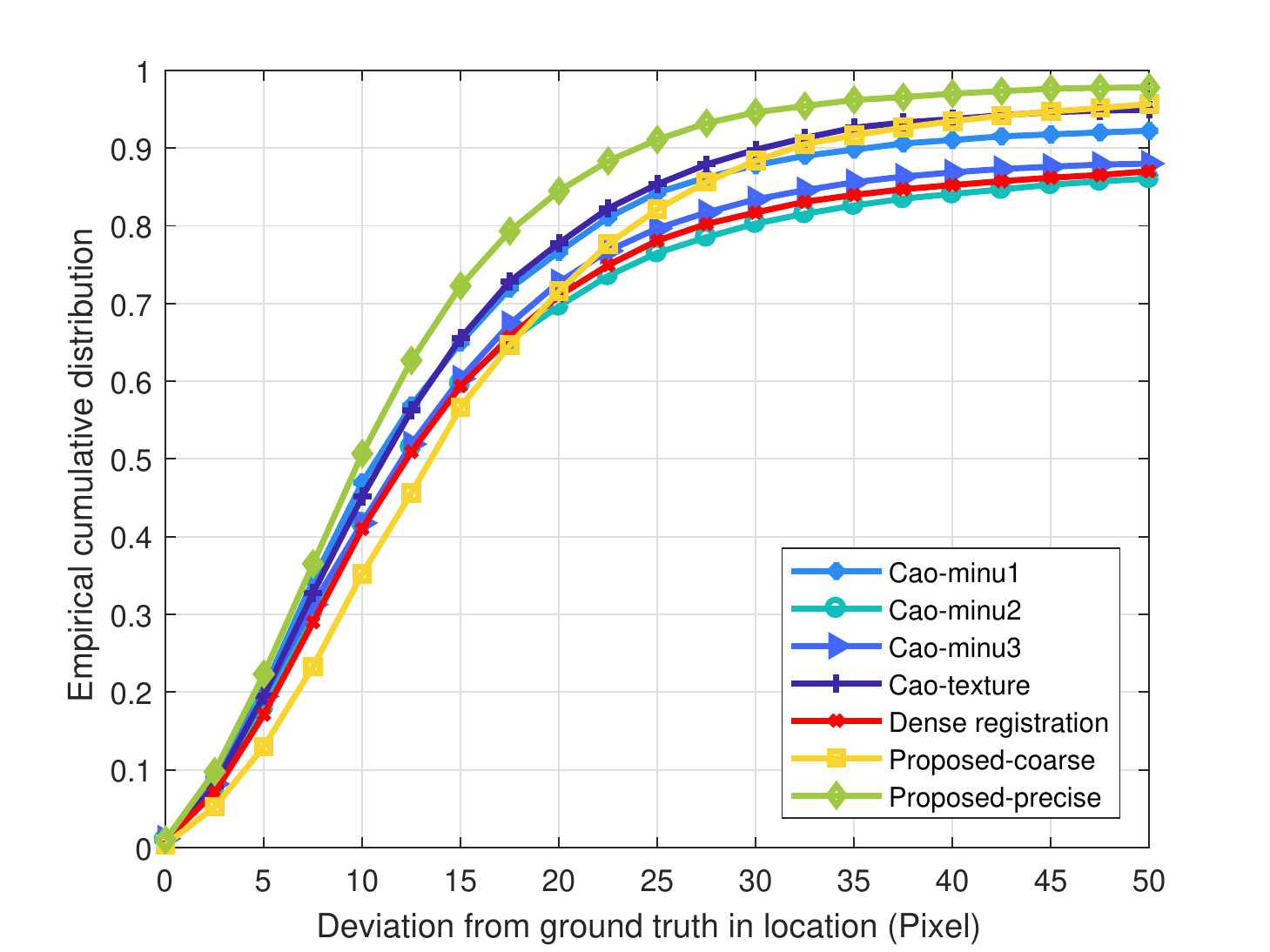}}
	\subfigure[]{\includegraphics[width=0.45\linewidth]{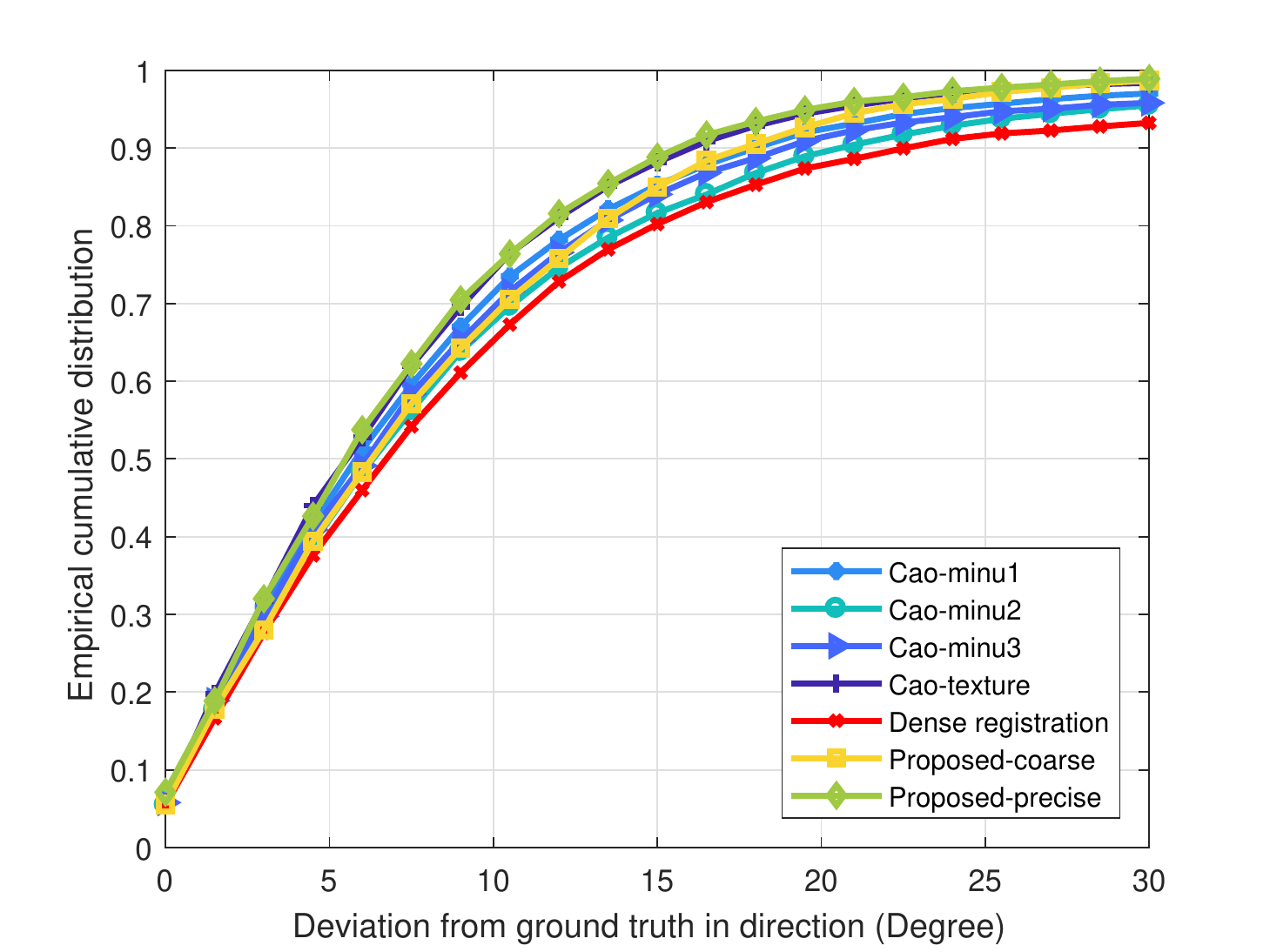}}
\end{center}
	\caption{Empirical cumulative distribution functions of (a) location and (b) direction  deviations on all the 258 pairs of fingerprints in NIST27 dataset.
}
\label{fig: matchperformanceall}
\end{figure*}

\textbf{Generation of Key Point Patches}
By adopting the proposed fingerprint simulation method,  twenty more fingerprints are simulated from a rolled fingerprint, such that more expressions are available for each fingerprint.  After that, key point  patches are generated to train the descriptor network.

We construct two training datasets to train the descriptors where image patches  are centered on different key points, the minutiae and the sampling points.
The minutiae based dataset is used to train a base model
and the sampling point based dataset is further applied to fine-tune the base model in order to make the model more generalizable to test data.

In both training datasets, the generation of images patches is the same. We take the minutiae based dataset as an example.
 For original rolled and latent fingerprints, as described earlier, minutiae are extracted from original fingerprint images,  and minutiae pairs have been obtained from the registration result. For the simulated fingerprints,  minutiae are estimated from that in rolled fingerprints with the distortion field, and the  correspondences between minutiae in different impressions can be obtained directly.
We cut minutiae patches of $200 \times 200$ pixels around each minutia  in the rolled, latent,  and simulated fingerprints separately for the minutiae descriptor training. 
Patches whose foreground area is less than 40\% of the patch area area excluded.
Based on the correspondences of minutiae from different impressions, the minutiae patches representing the same minutia share the same label. To ensure the correctness of each minutiae cluster,  only minutiae clusters with more than eight patches are used to train minutiae descriptors.

The above patch generation algorithm can ensure that the image patches with the same label come from the same key point. To train the Siamese network, we select a mini-batch which is composed of several positive pairs. These pairs with a same label make up of the positive set, and negative set is sampled randomly from different  categories.

\subsection{Global Patch Matching}

After obtaining the spatial transformation  parameters and similarities between all pairs of sampling points, we first select the pairs whose similarity is above a given threshold  $\tau$ as potential sampling point correspondences, and then use the spectral clustering based global patch matching method to compute the compatible subset of correspondences.

Different from the spectral clustering based registration in minutiae based approaches \cite{leordeanu2005spectral}, not only similarities but also the relative transformation parameters between pairs of sampling points are used in our method.  
For one pair of sampling points, we adjust the location and direction of sampling point on the latent fingerprint  based on the estimated transformation.  In this way, for different sampling points on the rolling fingerprint, the position and direction of the same point on the latent fingerprint are different after adjustment.

With the adjusted sampling point, the following registration is the same with that in minutiae based approaches. Specially, in consideration of time efficiency and accuracy, we choose mutually $N$ neighbors from all pairs of adjusted sampling points as initial correspondences based on their similarities. The descriptor similarities are normalized by min-max normalization. In the following, the second-order graph matching \cite{fu2013extended} is applied to  remove false correspondences.

\subsection{Efficiency Improvement}
 
The proposed method is very time-consuming in the online stage because it is required to estimate the relative transformation parameters for each pair of image patches,   then extract the descriptors of aligned patches,  and finally compute their similarities.  
The time required for a pair of fingerprints is equal to the time required for a pair of image patches multiplied by the number of matches.
The key to improve the  efficiency of the proposed method  is to  reduce the time required for a pair of patches.

We take the following strategy to  convert several steps that were computed online to offline. Concretely, we extract descriptors of dense sampling points on rolled and latent fingerprints  in the offline stage and store as the feature templates.  Then the descriptor of one sampling point in the online stage is approximated by the descriptor of its nearest neighbor in the template.  

When creating the   feature templates, the sampling points  should be  as dense as possible. For rolled fingerprints, the sampling interval is the same with that in the  registration method ($80 \times 80$), and the local orientation is regarded as the direction of sampling point.  For latent fingerprints,  we use $16 \times 16$ sampling points and give each sampling point 19 possible directions, which is sampled every 10 degrees between -90 degrees and 90 degrees.

It should be noted that the location and direction of sampling points have a certain error when extracting descriptors using this strategy.  But due to the very dense sampling point on latent fingerprints,  the maximum error is within 8 pixels in location and 5 degrees in direction.  The current descriptor extraction method is robust to such degree of location and direction error.

\begin{table*}
\caption{Registration accuracy with the thresholds of location deviation from ground truth as 20 pixels and direction deviation as 15 degrees.}
\label{tab:accuracy}
\begin{center}
\renewcommand\arraystretch{1.2}
\begin{tabular}{lccccccccc}
\toprule
&\multicolumn{4}{c}{Location} & \multicolumn{4}{c}{Direction}  \\
\cmidrule(lr){2-5}  \cmidrule(lr){6-9}  
 Method &  Good   & Bad  & Ugly  & All &  Good   & Bad   & Ugly  & All\\
\midrule
Cao-minu1       & 84.54\% &74.50\% & 62.29\%&76.65\%&89.35\%&82.28\%&80.00\%&85.23\%\\
Cao-minu2       & 81.22\% &66.72\% & 48.43\%&69.65\%&88.42\%&77.09\%&72.66\%&81.64\%\\
Cao-minu3       & 83.94\% &66.64\% & 56.16\%&72.74\%&89.05\%&80.80\%&77.25\%&84.04\%\\
Cao-texture     & 83.13\% &77.84\% & 66.06\%&77.75\%&\textbf{90.97\%}&85.47\%&85.41\%&88.16\%\\
Dense registration  & 80.79\% &66.79\% & 55.14\%&71.00\%&86.75\%&76.87\%&70.28\%&80.22\%\\
Proposed-coarse & 80.20\% &69.38\% & 55.69\%&71.57\%&88.97\%&84.14\%&77.34\%&84.96\%\\
Proposed-precise&\textbf{87.22\%} &\textbf{84.51\%} & \textbf{78.53\%}&\textbf{84.48\%}&90.63\%&\textbf{85.92\%}&\textbf{88.53\%}&\textbf{88.82\%}\\
\bottomrule
\end{tabular}
\end{center}
\end{table*}

\subsection{Implementation Details}

A database from local police department is used to construct the training data in both local patch alignment and matching procedures. This dataset contains more than one thousand pairs of rolled and latent fingerprints, where latent fingerprints are collected from real crime scenes. 

We choose 500 pairs of mated latent and rolled fingerprints in the database and exclude those whose minutia matching scores computed by VeriFinger \cite{NeuroTech} are less than 50. Fingerprints of relatively good quality were chosen so that the data simulation results for training deep neural networks are reliable. FingerNet \cite{tang2017fingernet} is adopted to extract the ROI, minutiae, and ridge orientation images of both rolled and latent fingerprints. 

All the networks in our method are trained with PyTorch  \cite{paszke2017automatic} from scratch on NVIDIA GTX 1080 Ti GPUs. 
The stochastic gradient descent (SGD) is used for optimization with a weight decay of $5\times10^{-4}$. 

To balance the loss function, we use $\lambda_{match}=0.25$, and $\lambda_{simi}=0.5$.

\textbf{Local Patch Alignment} 
We train two alignment models with different synthetic training datasets for two registration stages, both with learning rate $10^{-3}$, weight decay 0.95 every 5 epochs, and batch size of 32. Models are all trained for over 20 epochs.

\textbf{Local Patch Matching}
The descriptor network is trained with the mini-batch which contains multiple pairs of patches with the same label, and on-line hard negative sampling is used to select the negative pairs with smallest distance during training for better convergence and accuracy.

When training the base model, the learning rate is set initially as $10^{-2}$ and decays at 0.95 rate every 5 epochs. The batch size is set as 32, and the model is trained for over 20 epochs. When fine-tuning the model with sampling point based training data, we use the initial learning rate $10^{-4}$, and other parameters are the same. 

\textbf{Global Patch Matching}
In coarse registration stage, we choose mutually $N=4$ neighbors as initial matching pairs, and the threshold $\tau=0.5$ to exclude obvious dissimilar point pairs. In precise registration,  $N$ is set as $4$, and  $\tau$ is set as $0$.

\begin{figure*}
\begin{center}
	\subfigure[]{\includegraphics[width=0.32\linewidth]{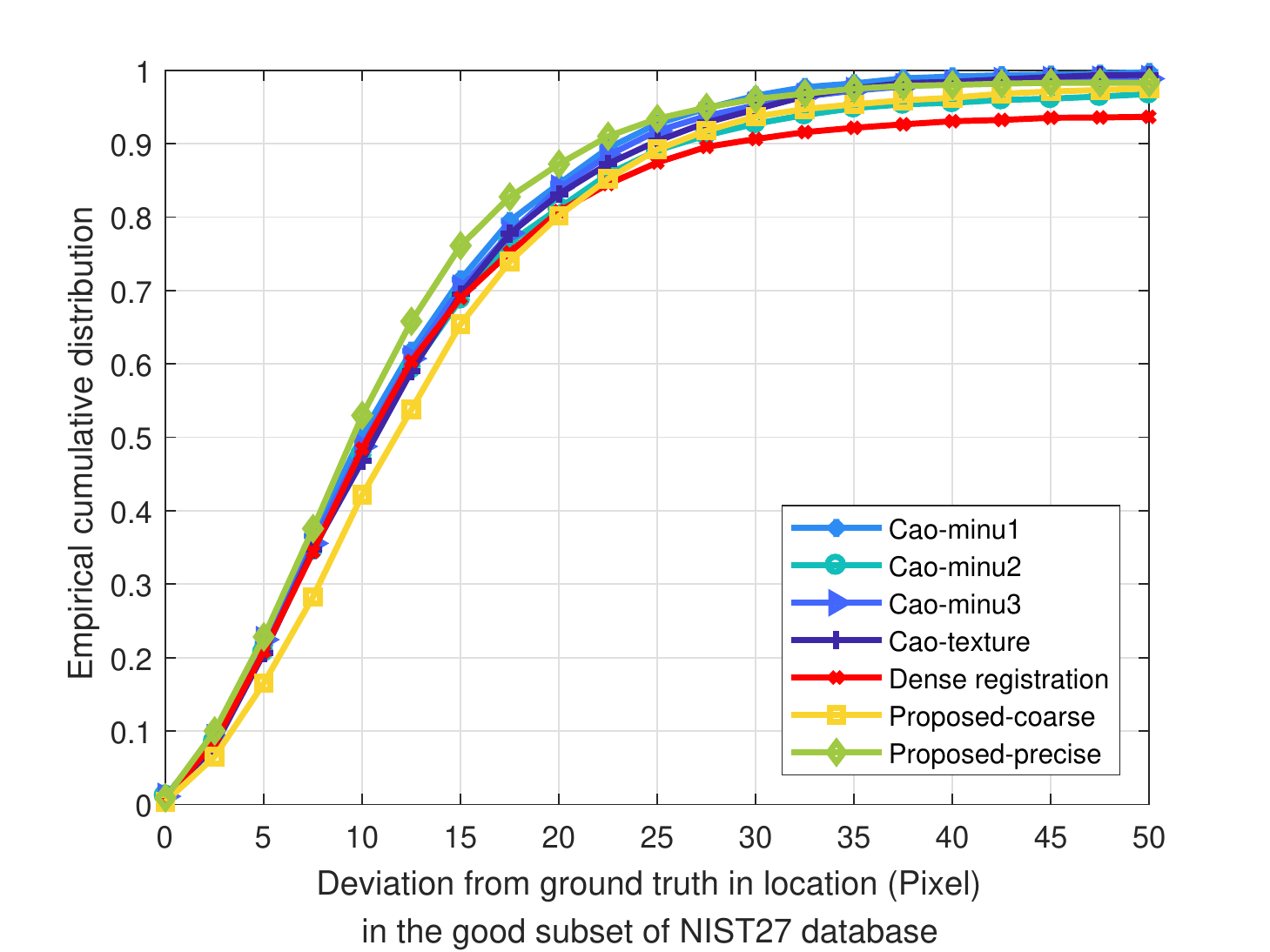}}
	\subfigure[]{\includegraphics[width=0.32\linewidth]{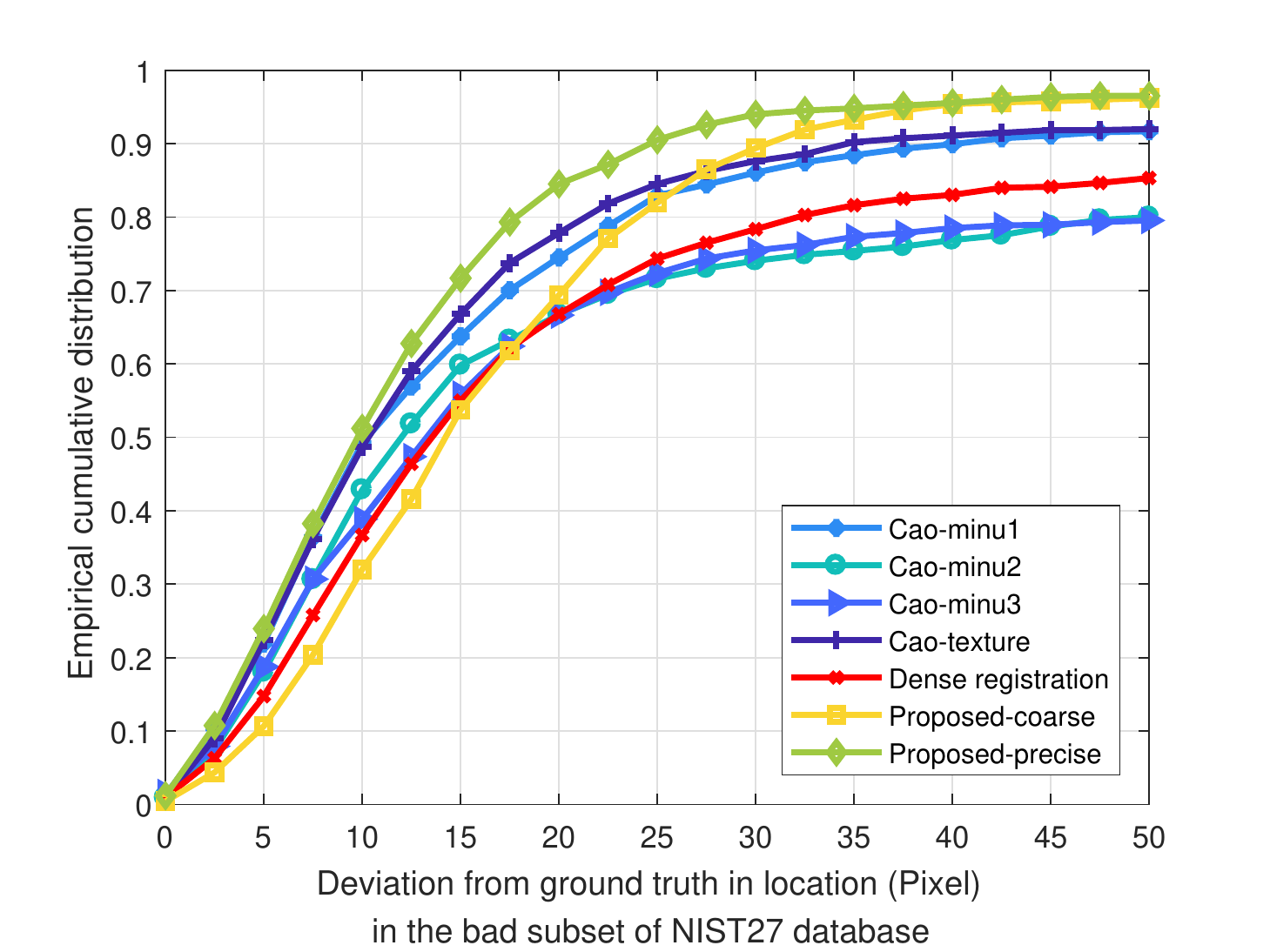}}
	\subfigure[]{\includegraphics[width=0.32\linewidth]{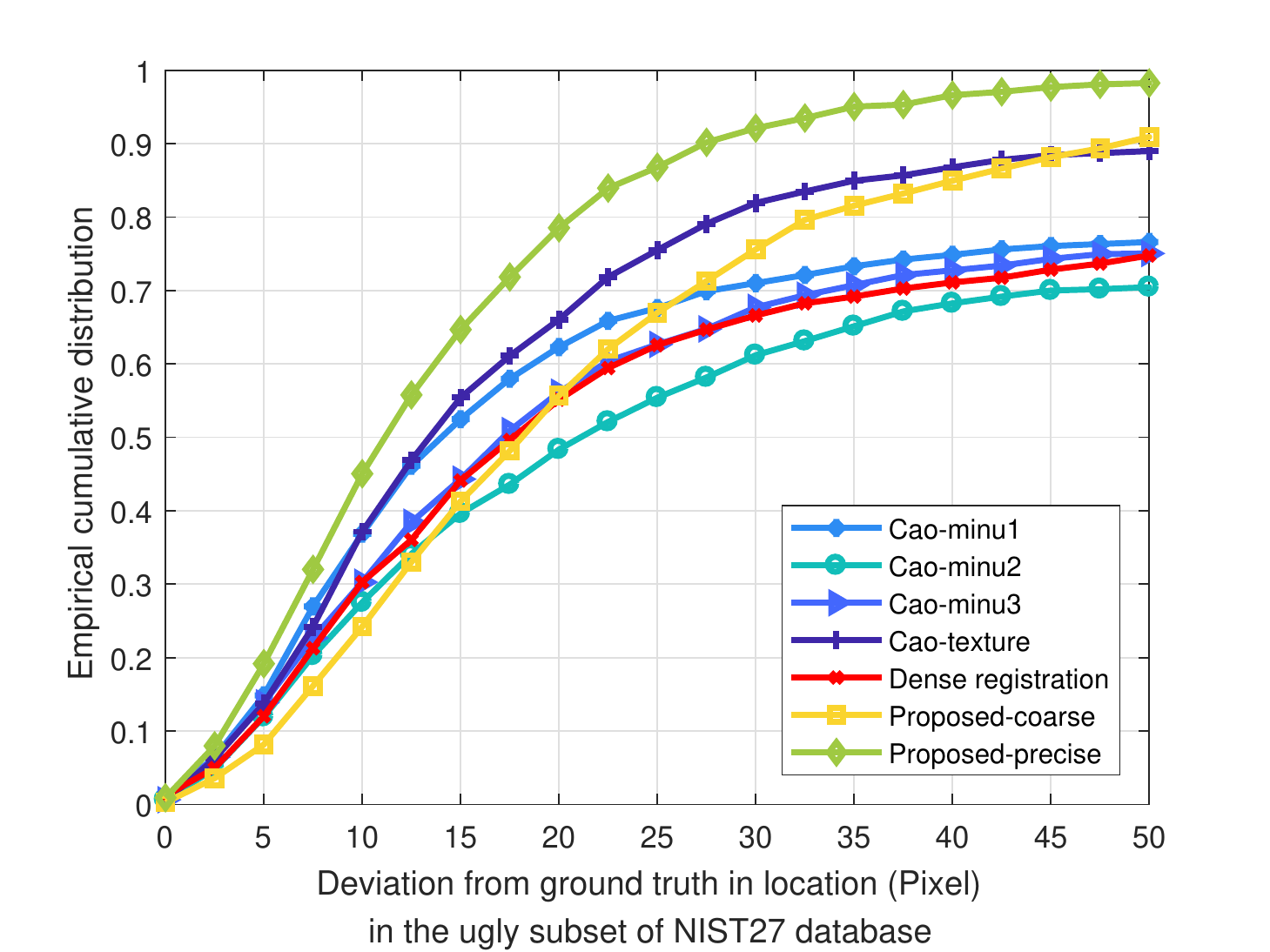}}
	\subfigure[]{\includegraphics[width=0.32\linewidth]{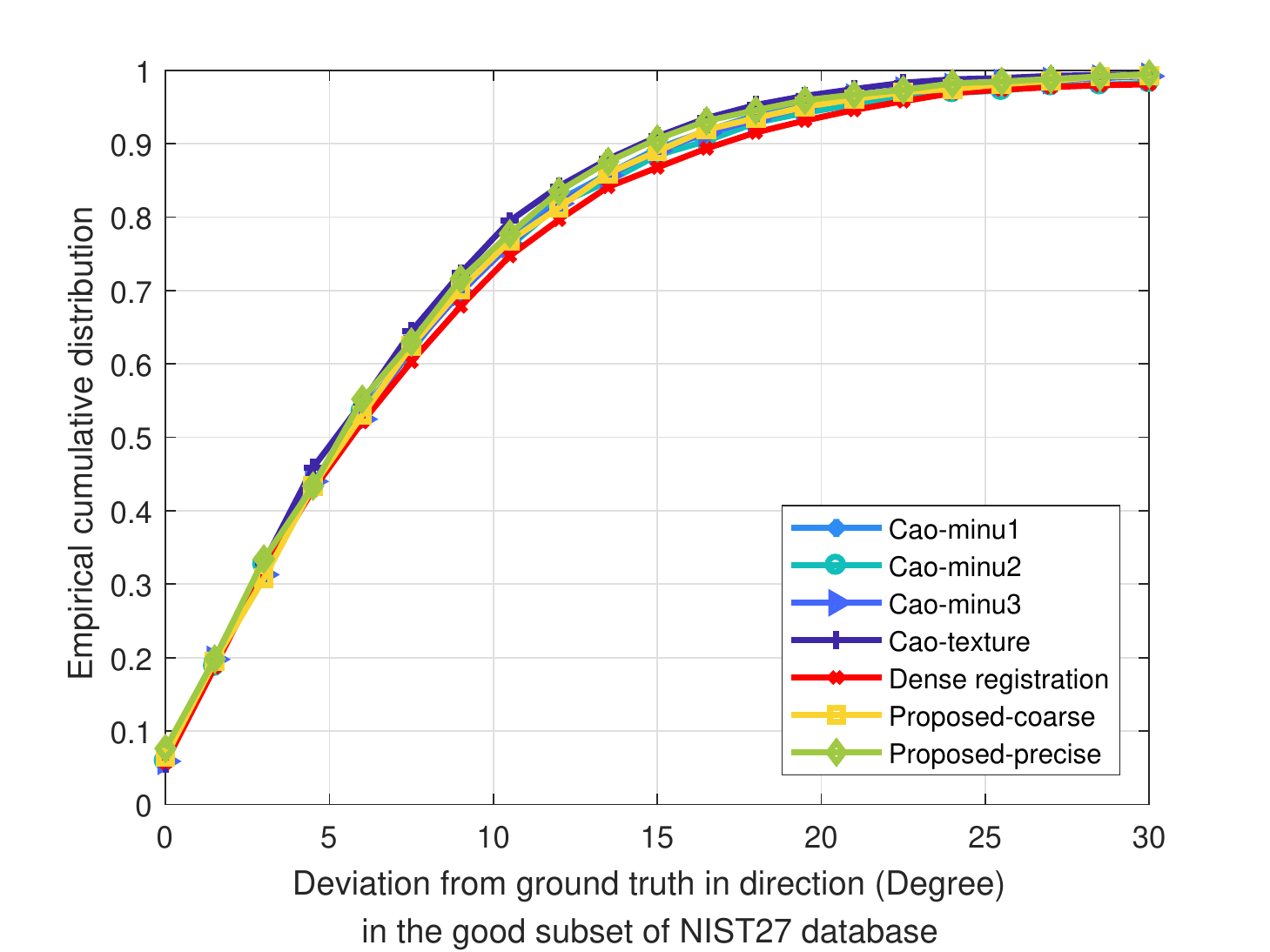}}	
	\subfigure[]{\includegraphics[width=0.32\linewidth]{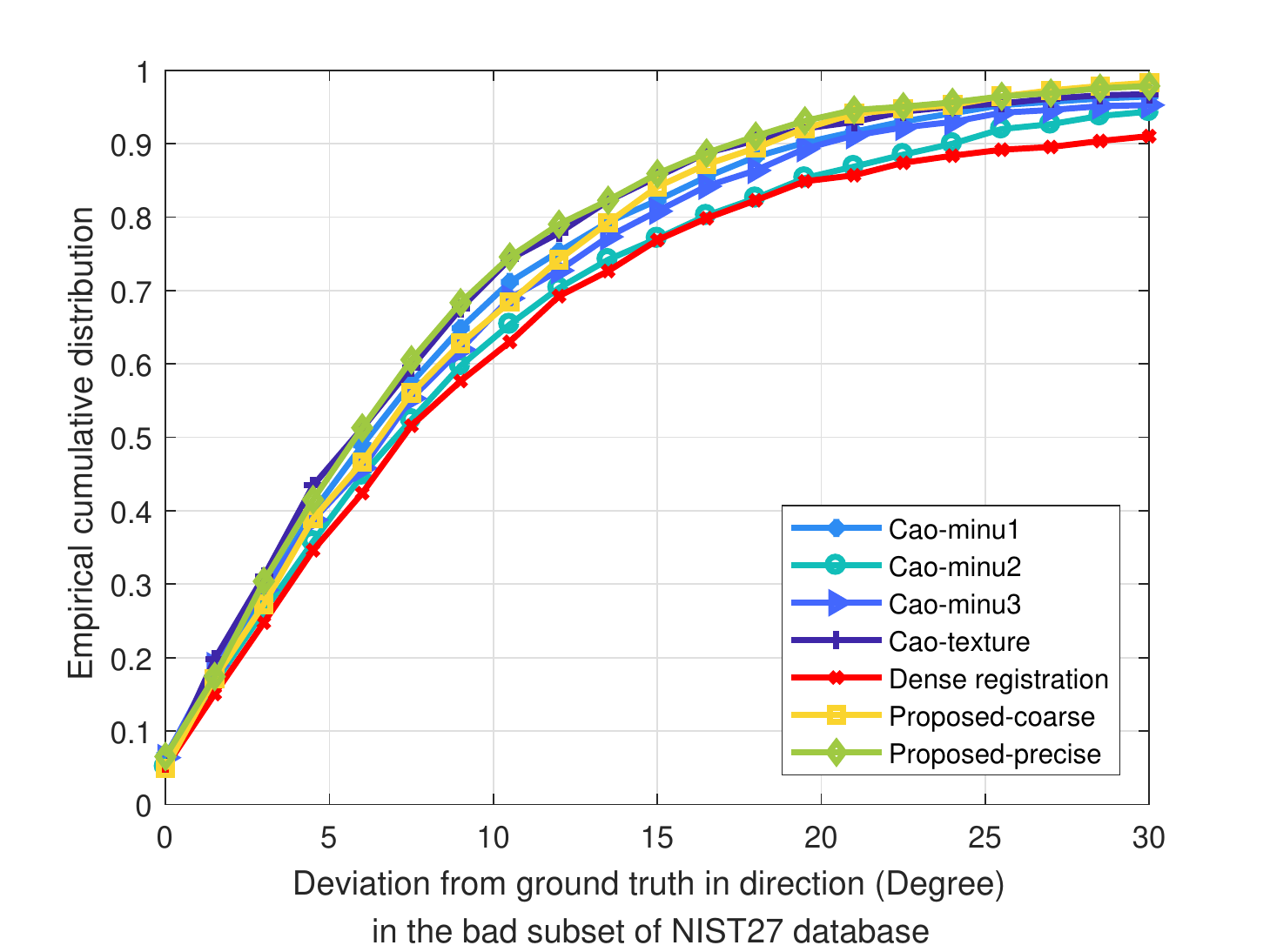}}	
	\subfigure[]{\includegraphics[width=0.32\linewidth]{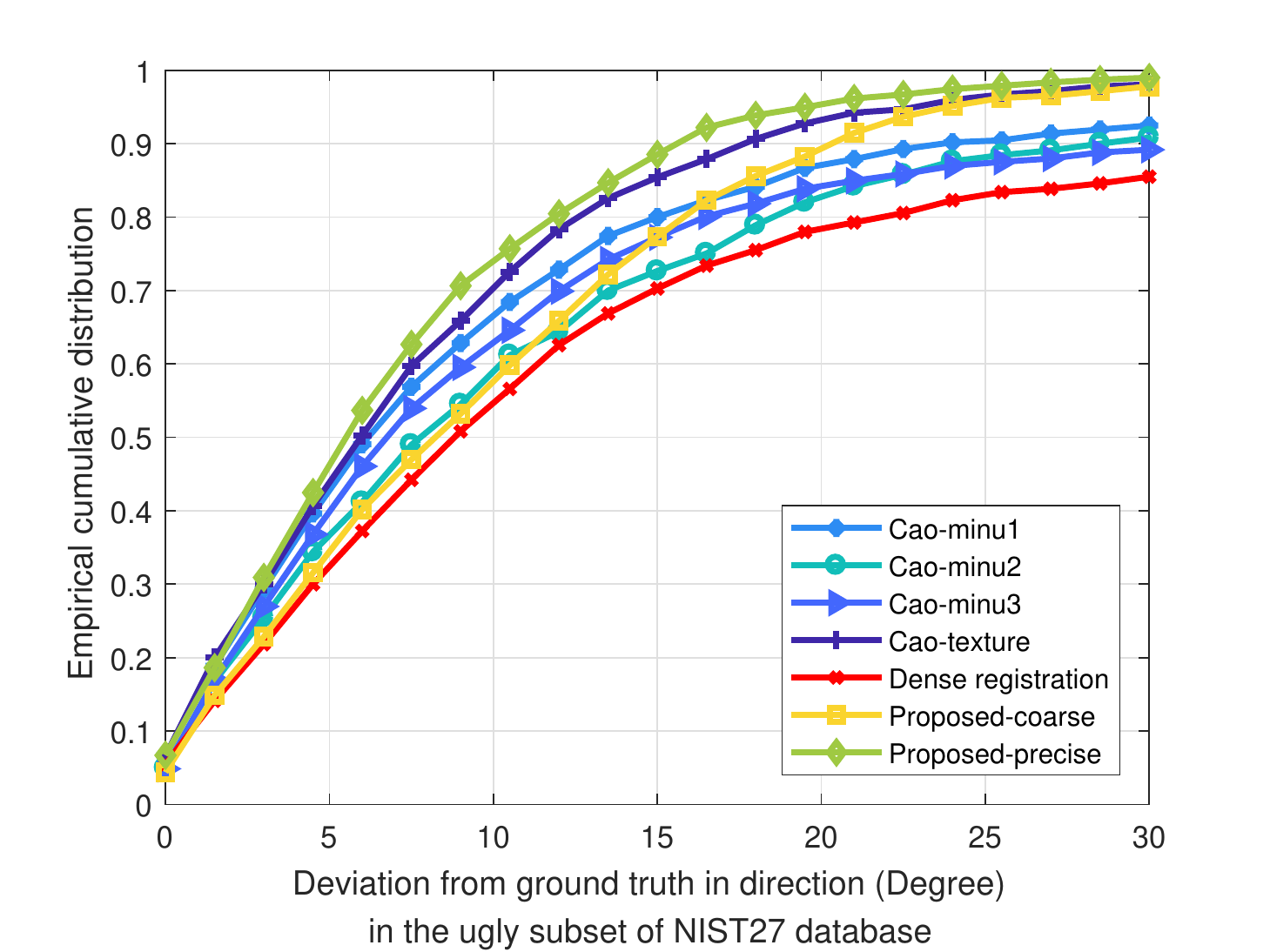}}
\end{center}
	\caption{Empirical cumulative distribution functions of (a)(b)(c)  location and (d)(e)(f) direction  deviations on the good, bad, and ugly subsets of fingerprints in NIST27 dataset, respectively.}
\label{fig: matchperformance}
\end{figure*}

\begin{figure*}[t]
\begin{center}
	\includegraphics[width=0.85\linewidth]{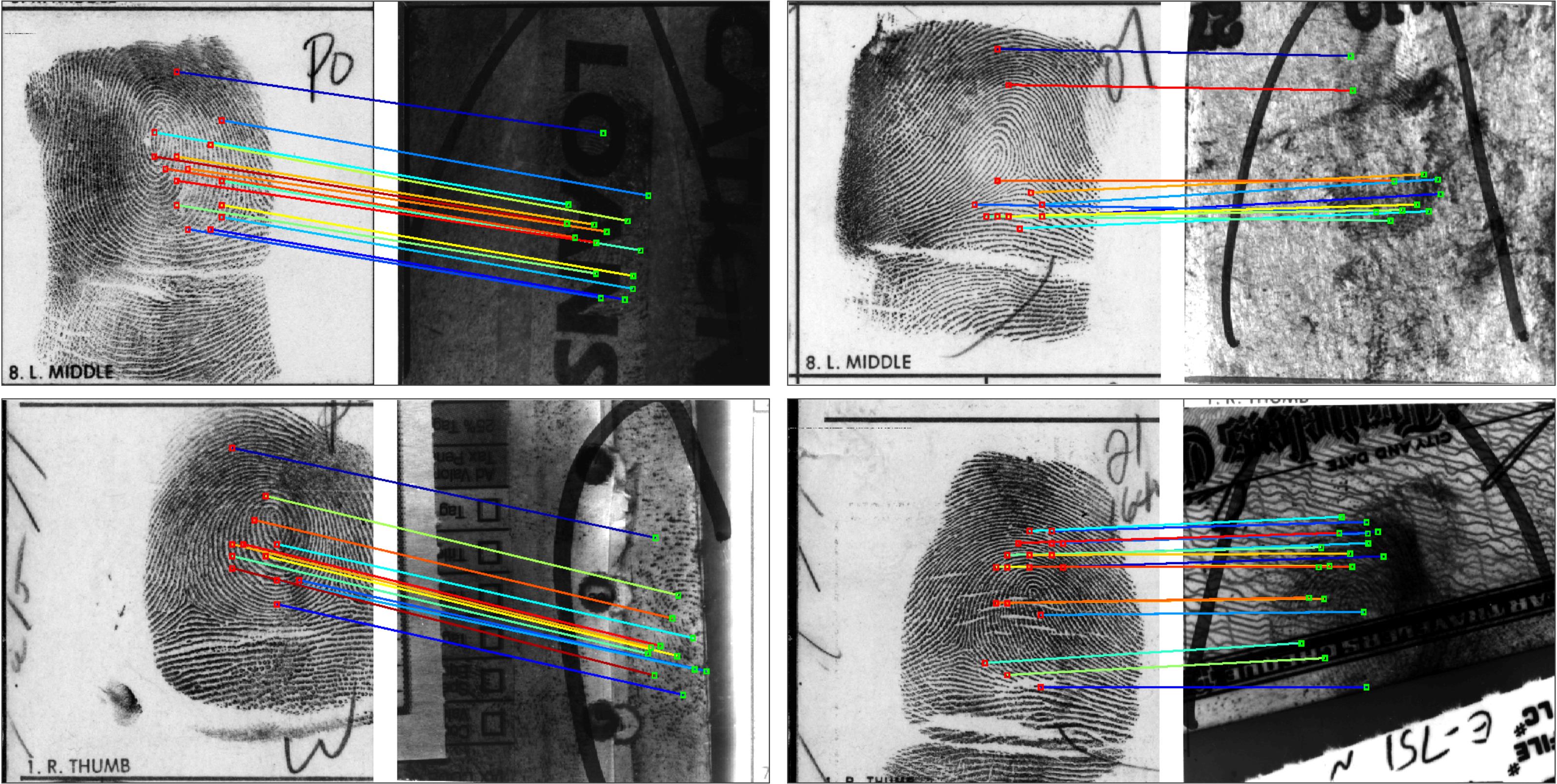}
\end{center}
	\caption{ Examples of correctly registered fingerprints pairs in the NIST27 database. }
\label{fig: goodexample}
\end{figure*}

\section{Experiments}

In this section, we carry out several experiments to  evaluate the performance of the proposed latent fingerprint registration algorithm and compare it to   the dense registration \cite{si2017dense} and Cao's method \cite{cao2018end}, which show  great performance on slap fingerprint registration and latent fingerprint recognition, respectively. The experiments are performed on the NIST27 database and MOLF database. 

Firstly, we directly evaluate the registration accuracy on NIST27 database by comparing the deviation between mated minutiae after aligning the mated fingerprints with the estimated registration parameters. Secondly, we use a matching experiment on both databases as an indirect evaluation considering that the purpose of fingerprint registration is to improve the accuracy of fingerprint recognition, and there are no  ground-truth registration parameters in MOLF database.

\subsection{Databases}

\textbf{NIST27 database}  \cite{NIST27} is a public latent fingerprint database containing 258 pairs of latent and rolled fingerprints. For a fingerprint image in the database, we use manually  marked ROI of latent fingerprints \cite{yoon2011latent} and use FingerNet \cite{tang2017fingernet} to obtain the ROI of rolled ones. In order to guarantee the fairness, the same ROI is applied to all algorithms to be compared.

\textbf{MOLF database} \cite{sankaran2015multisensor} contains 19,200 fingerprints from 100 subjects with five different capture methods.  We choose the DB3$\_$A subset as the reference dataset, which is captured using CrossMatch L-Scan Patrol, and the DB4 subset, which is the only latent fingerprint dataset. For each fingerprint, the first instance is selected. Finally, 1,000 pairs of latent and plain fingerprints from 10 fingers of 100 users are obtained. We use the first 500 pairs of fingerprints to conduct the experiments. 
Considering  the clean fingerprint background in these two datasets,  the Otsu's method \cite{otsu1979threshold} and morphological operations are used to obtain the ROI of each fingerprint.

\subsection{Performance of Fingerprint Registration Evaluated by Mated Minutiae Pairs}

We use the location and direction differences of ground truth matching minutiae after alignment as the evaluation metric. In the NIST27 database, each pair of mated rolled and latent fingerprints has matching minutiae provided by fingerprint experts. 
After obtaining sampling point correspondences, the translation and rotation between image pairs are computed by the average translation and rotation between sampling point pairs. With the spatial transformation parameters, latent fingerprints can be aligned to rolled ones, and the manually marked minutiae on latent fingerprints can also be registered.

Since Cao \textit{et al.} constructed three different minutiae templates and one texture template, and outputted one fingerprint registration result for each template,   the comparison is conducted with their four registration  results.

The cumulative distribution functions of location and direction deviations are shown in Fig. \ref{fig: matchperformanceall}. As we can see

\begin{itemize}
\item  The performance of the dense registration is comparable with that of minutiae based methods such as the the first and second template in Cao's method, probably because it employs minutiae based initial registration.
\item The performance of the proposed precise registration method is more accurate than that of coarse registration. After the precise registration, more than 80\% minutiae have the deviation less than 20 pixels in location and  15 degrees in direction.
\item The first minutiae template in \cite{cao2018end}  achieves better results than other two minutiae templates, and the performance of texture template is slightly better.  
 Compared with other methods, the performance of ours is consistently better.   
\end{itemize}

For concrete analysis, we compare the performance on three subsets of NIST27 database, which are named as good, bad, and ugly subsets. The comparison results are shown in Fig. \ref{fig: matchperformance}. Besides, to better analyze the results, we list the registration accuracy of different methods on three subsets and the whole dataset  in Table \ref{tab:accuracy} when setting the threshold of deviation from ground truth as 20 pixels and  15 degrees in location and direction, respectively. 
We can observe that 
\begin{itemize}
\item The proposed method achieves comparable results on the good subset and performs better on the bad and ugly subset. As shown in the table, our method can achieve 2.68\%,  6.67\%, and 12.47\% improvement in location in the good, bad, and ugly subset, 0.45\% and 3.12\% improvement in direction in the bad and ugly subset, respectively, in comparison with the best results of other methods. 
 The improvement  is greater when the image quality is worse, which suggests the superiority of the proposed method on latent fingerprints with low image quality.

\item Among the four templates used in \cite{cao2018end}, the texture template performs better on bad and ugly subset than three minutiae templates. Its superior performance indicates that dense sampling points are suitable key points in latent fingerprints of very poor quality.
\end{itemize}

\begin{figure}[t]
\begin{center}
	\includegraphics[width=\linewidth]{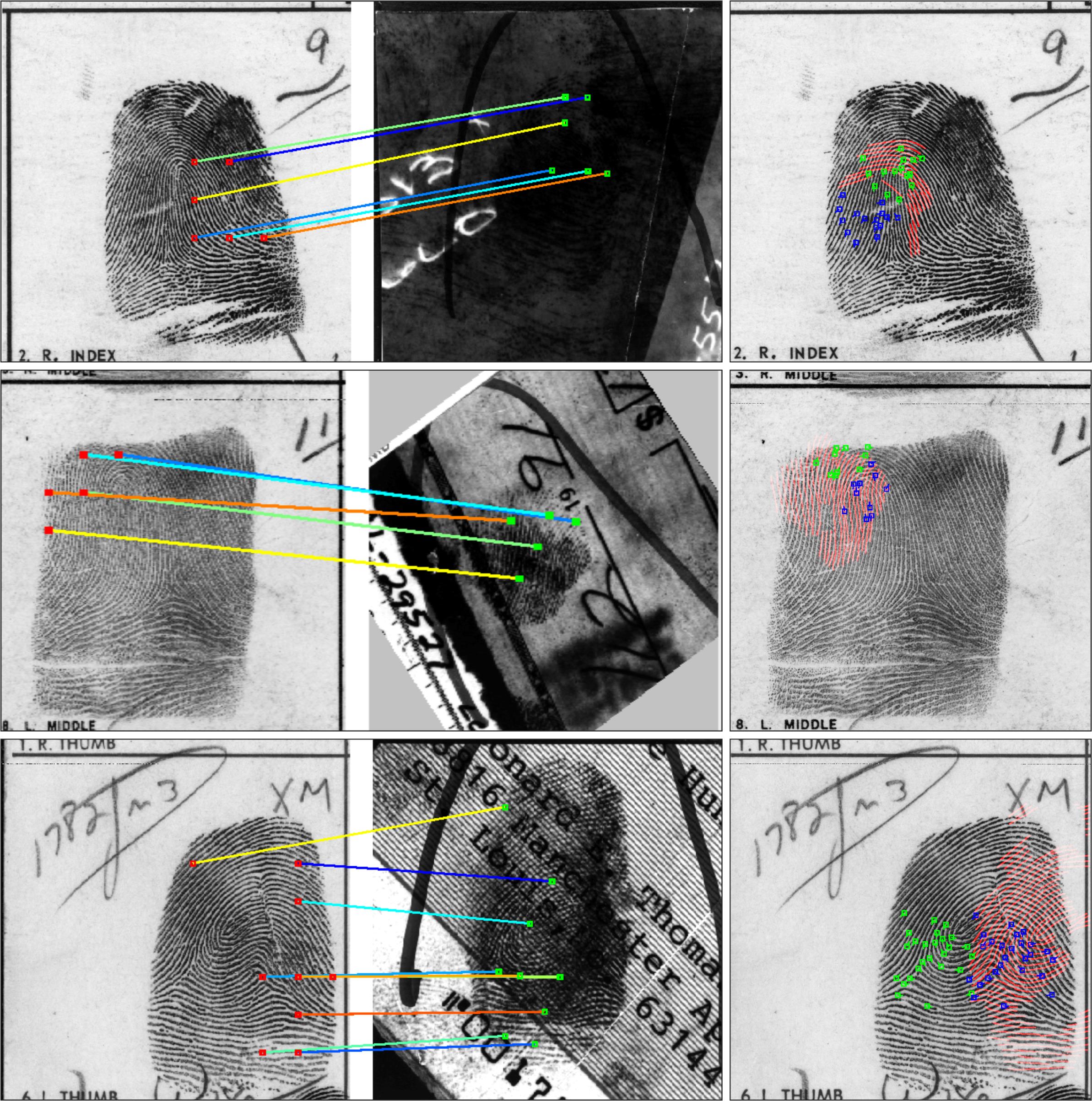}
\end{center}
	\caption{ Failure cases in the NIST27 database. Left two columns show the sampling point correspondences, and the third column shows the deviation between matching minutiae after coarse registration. 
 } 
\label{fig: badexample}
\end{figure}

Several examples of correct matches and failure cases are shown to discuss the superiority and inferiority of the proposed approach. 
Fig. \ref{fig: goodexample} illustrates several examples where correct matches can be found by the proposed method, but other methods are not successful. From these examples, it is demonstrated that the proposed method is robust in situations where fingerprint ridges are too deep or too light to distinguish from background or are polluted by background noise.
Fig. \ref{fig: badexample}  gives three failure cases.The failure is mainly due to limited fingerprint area and the great background noise. When the distances between matching points are too far after coarse registration, the effect of precise registration is limited. 
Therefore, here only shows the sampling point correspondences obtained by coarse registration.
These examples show that the proposed method needs further improvement to deal with fingerprints with very small effective area or  with very poor quality. 

\subsection{Performance of Fingerprint Registration Evaluated by Fingerprint Matching}

For each pair of input fingerprints,  we register the latent fingerprint to the rolled one using the registration approaches.  Then the deep descriptor proposed in \cite{cao2018end} is used to computing the similarity of two virtual minutiae from two fingerprints, and the average similarity of all virtual minutiae on their overlapping area is used as a matching score. The virtual minutiae are obtained by  sampling evenly at $24\times24$ intervals  with the local ridge direction of rolled fingerprints as their direction. 
If there is no registration parameters estimated,  the score is set to the minimum 0. 
After obtaining all the matching scores,  the Cumulative Match Characteristic (CMC) curve is used to evaluate the performance of the different registration algorithms.

\begin{figure}
\begin{center}
	\includegraphics[width=0.9\linewidth]{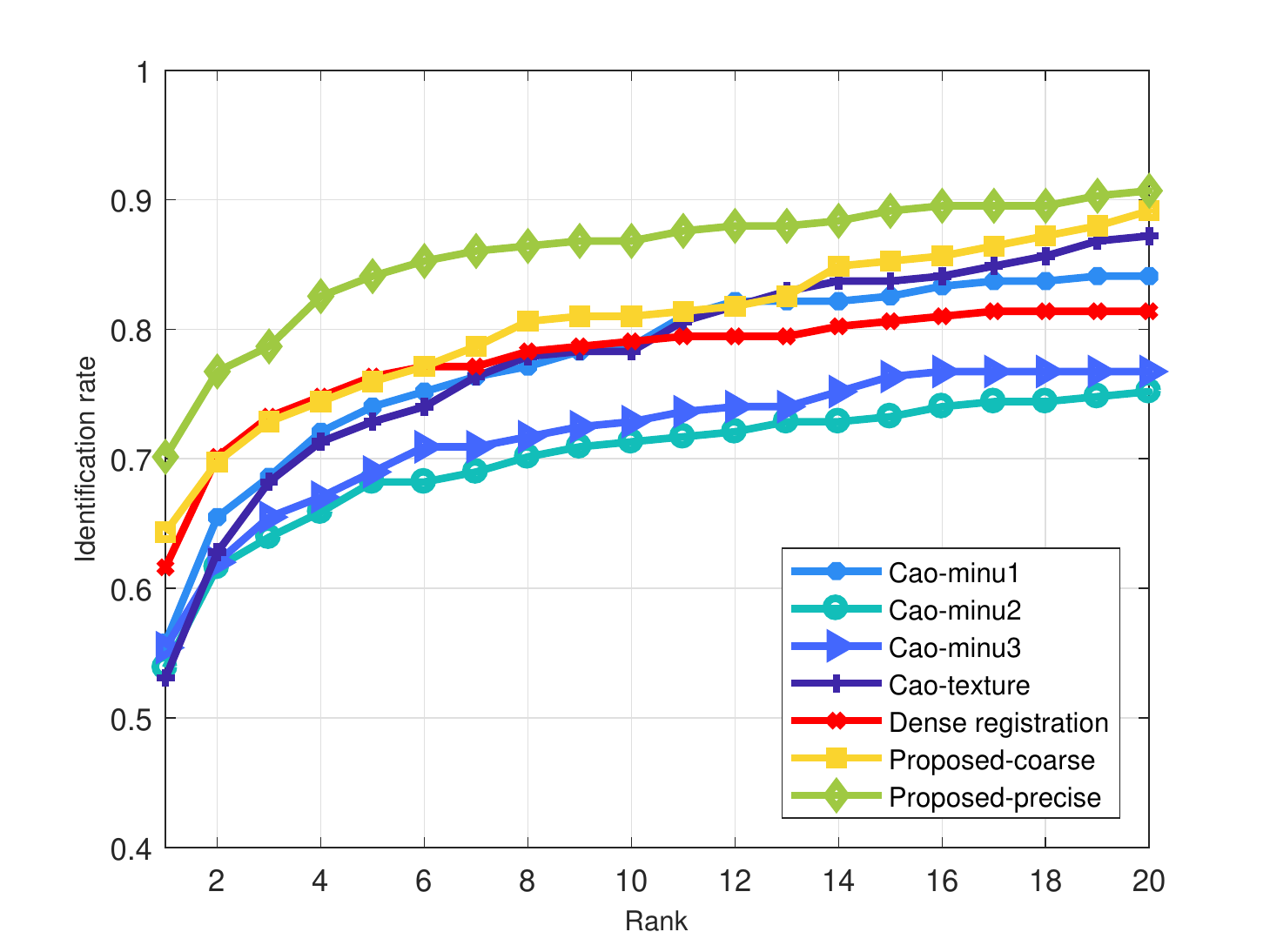}
\end{center}
	\caption{ The matching performance  of different registration methods on  NIST27 database.}
\label{fig: nist27matching}
\end{figure}

\begin{figure}
\begin{center}
	\includegraphics[width=0.9\linewidth]{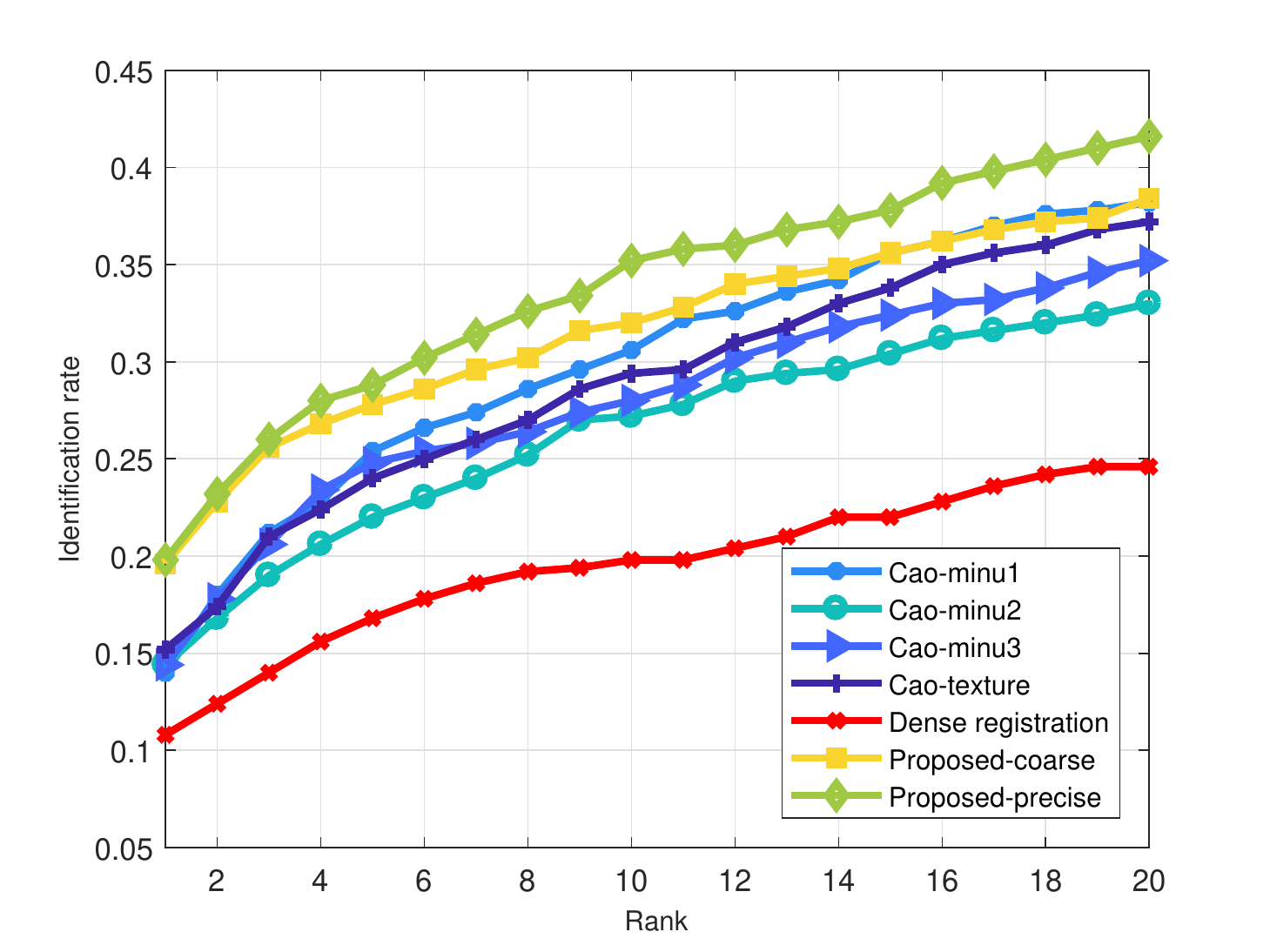}
\end{center}
	\caption{ The matching performance  of different registration methods on  the first 500 pairs of MOLF database.}
\label{fig: molfresult}
\end{figure}

 Fig. \ref{fig: nist27matching} shows the matching performance on NIST27 database.  The performance of dense registration algorithm is comparable to the best result of Cao's algorithm, but its identification rate on rank-1 is higher.  The proposed approach, especially the precise registration,  obviously outperforms other methods, where the  rank-1 identification performance  improves from 61.6\% to 70.1\%.
 The matching performance on MOLF database is shown in Fig. \ref{fig: molfresult}. This database is a rather difficult database since the rank-1 accuracy of all algorithms is very low, but the performance of our approach is still significantly improved (from 15.2\% to 19.8\% rank-1 accuracy).  This indicates that the proposed  algorithm   can be well generalized to different latent fingerprint databases. Specially, the performance of dense registration on this database is very poor,  probably because the fingerprint features it uses are extracted by traditional methods, which are more prone to errors.

\begin{figure}
\begin{center}
	\subfigure[Plain]{\includegraphics[width=0.23\linewidth]{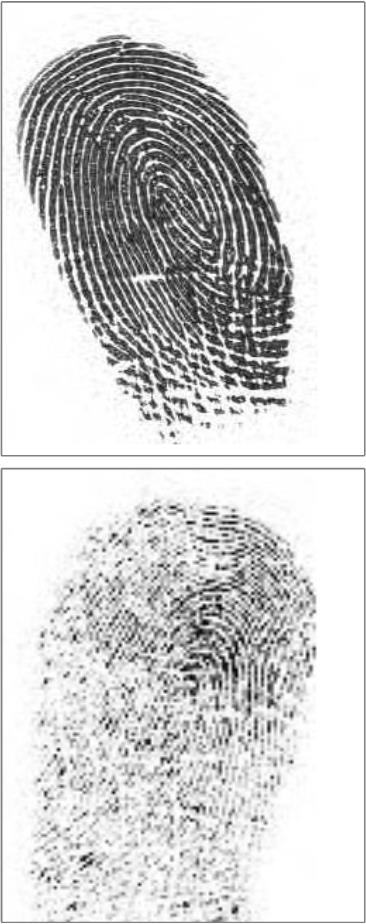}}
	\subfigure[Latent]{\includegraphics[width=0.23\linewidth]{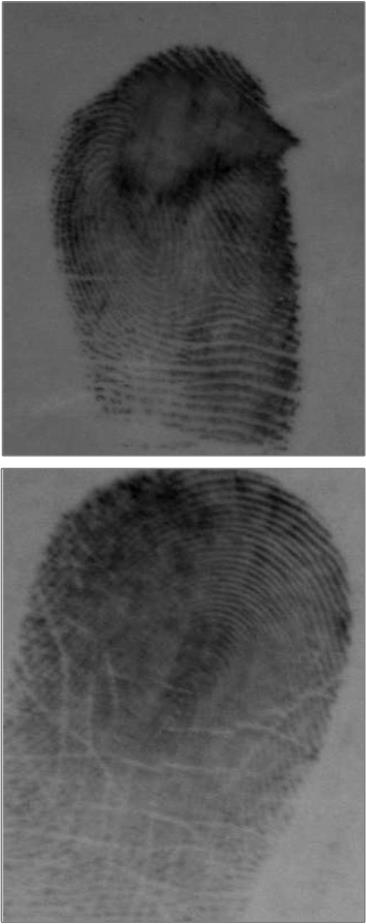}}
	\subfigure[Registered by the proposed method]{\includegraphics[width=0.23\linewidth]{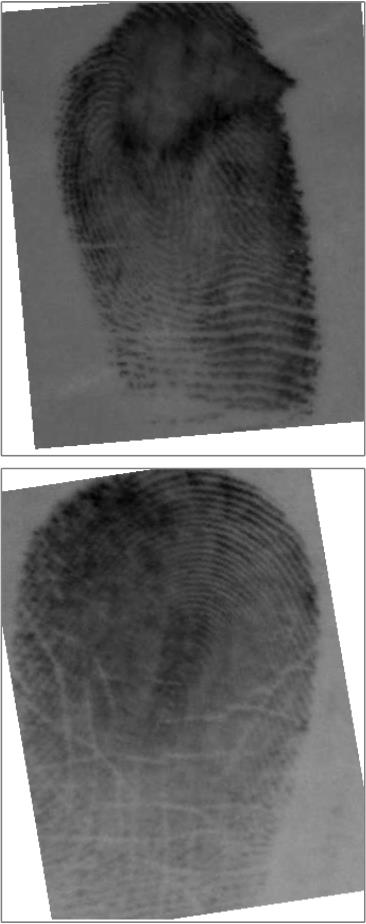}}
	\subfigure[Registered by method in \cite{cao2018end}]{\includegraphics[width=0.23\linewidth]{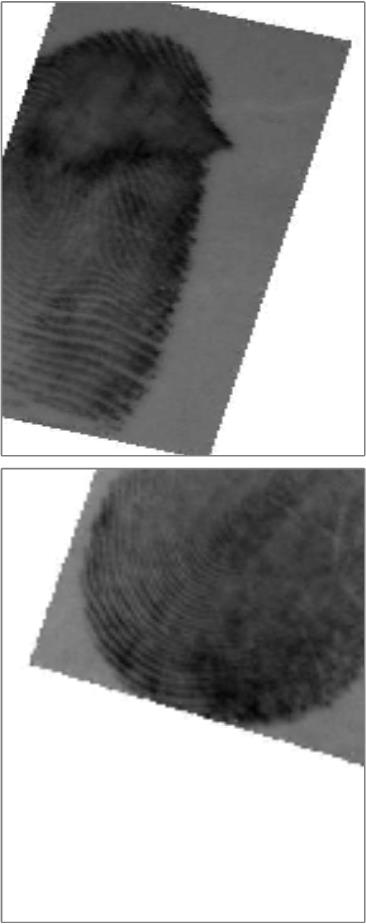}}
\end{center}
	\caption{Examples where the registration results of the proposed approach are more precise in the MOLF database.}
\label{fig: goodexamplemolf}
\end{figure}

 Fig. \ref{fig: goodexamplemolf} gives two examples on MOLF database where our method conducts the registration successfully. Only the registration results by the texture template in \cite{cao2018end} is shown for comparison since the minutiae based approaches (the other three minutiae templates and the dense registration) often fails. The ridge lines in these latent fingerprints are difficult to be accurately extracted by existing algorithms but can be identified by  human experts based on context information. The proposed method do not rely on estimated orientation field, and thus can handle such level of image blurry. 

\subsection{Effect of Coarse-to-Fine Registration}

\begin{figure}
\begin{center}	
	\subfigure[Latent to match]{\includegraphics[width=0.32\linewidth]{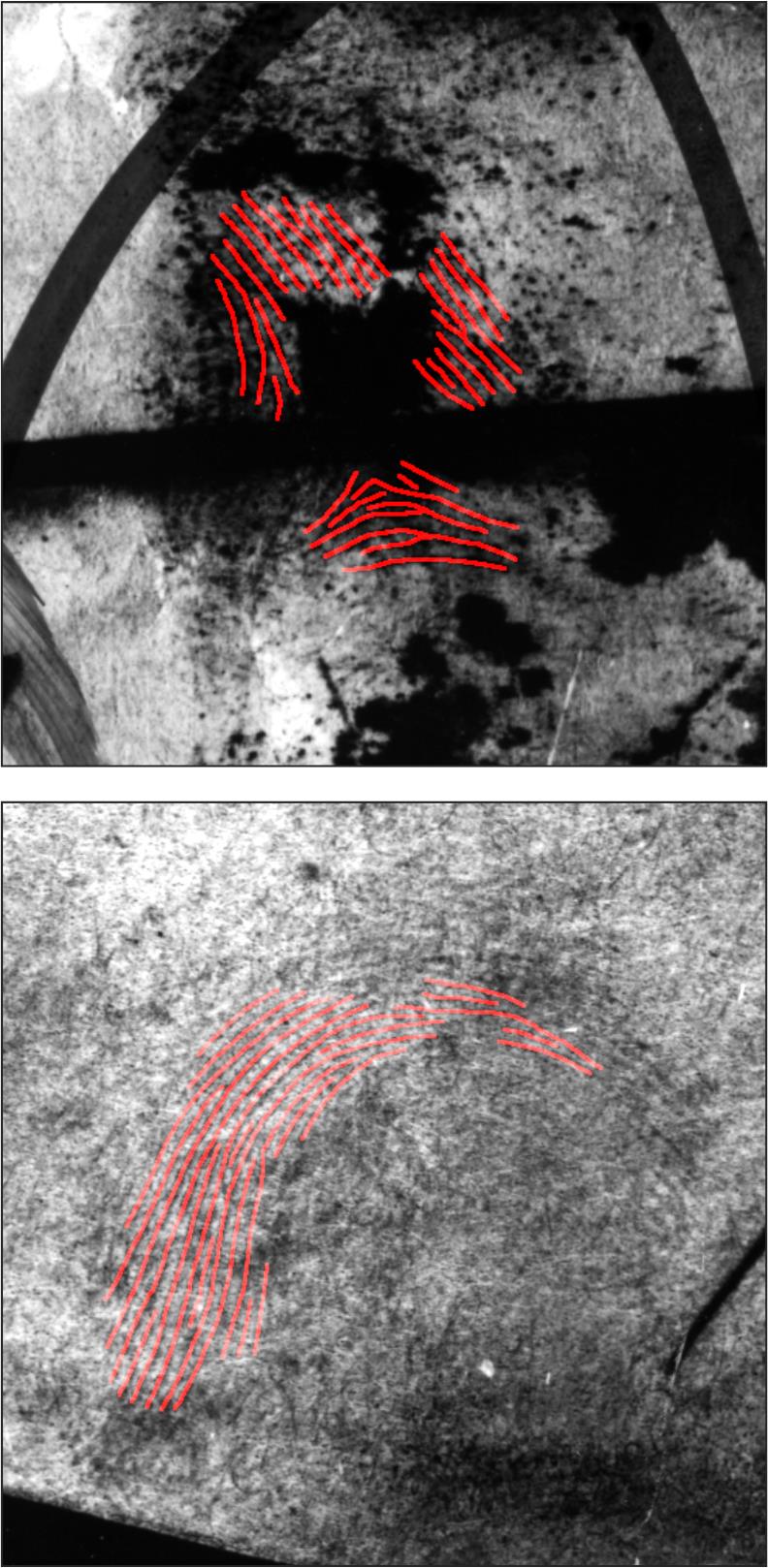}}
	\subfigure[Coarse matching]{\includegraphics[width=0.32\linewidth]{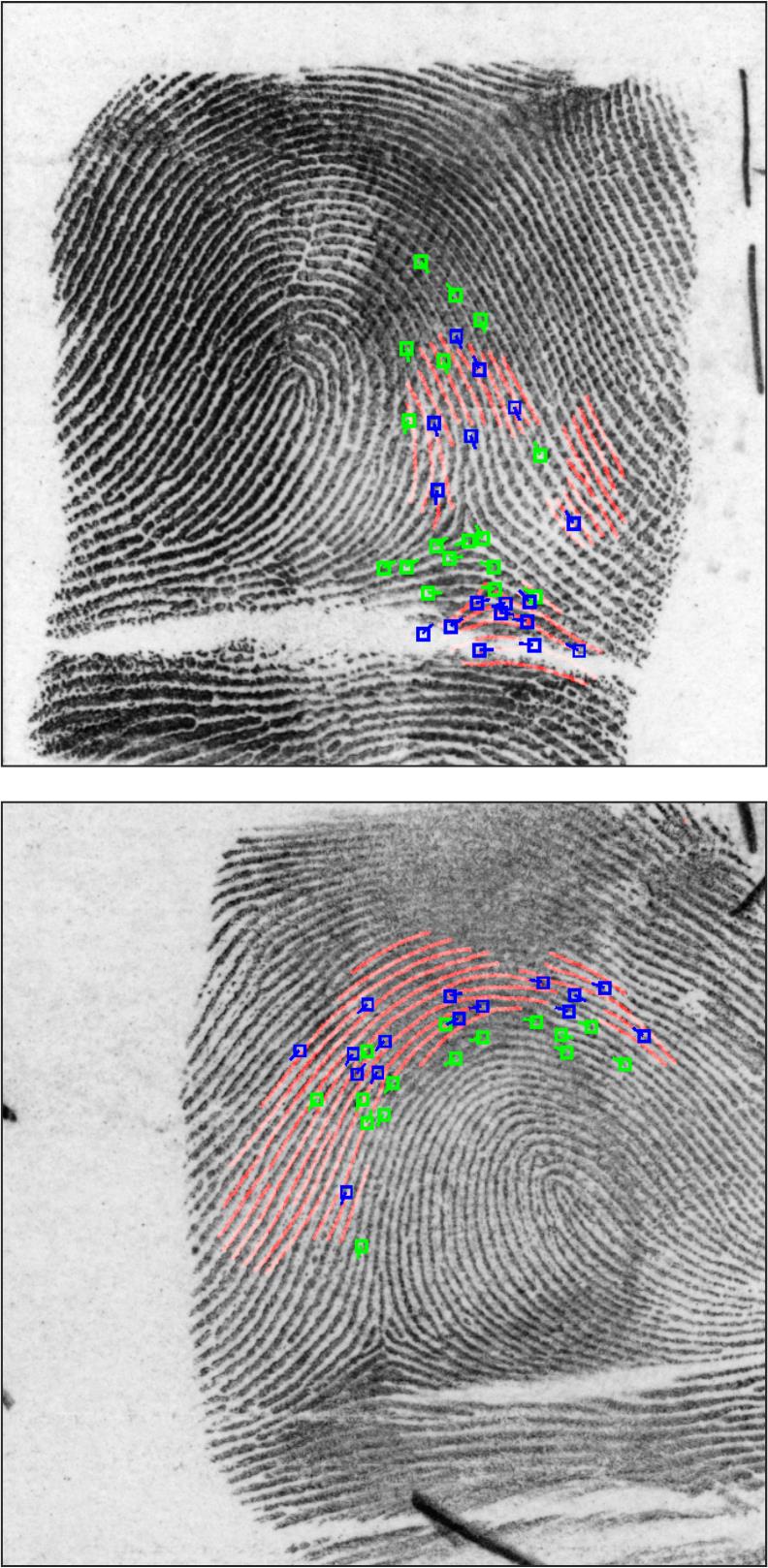}}
	\subfigure[Precise matching]{\includegraphics[width=0.32\linewidth]{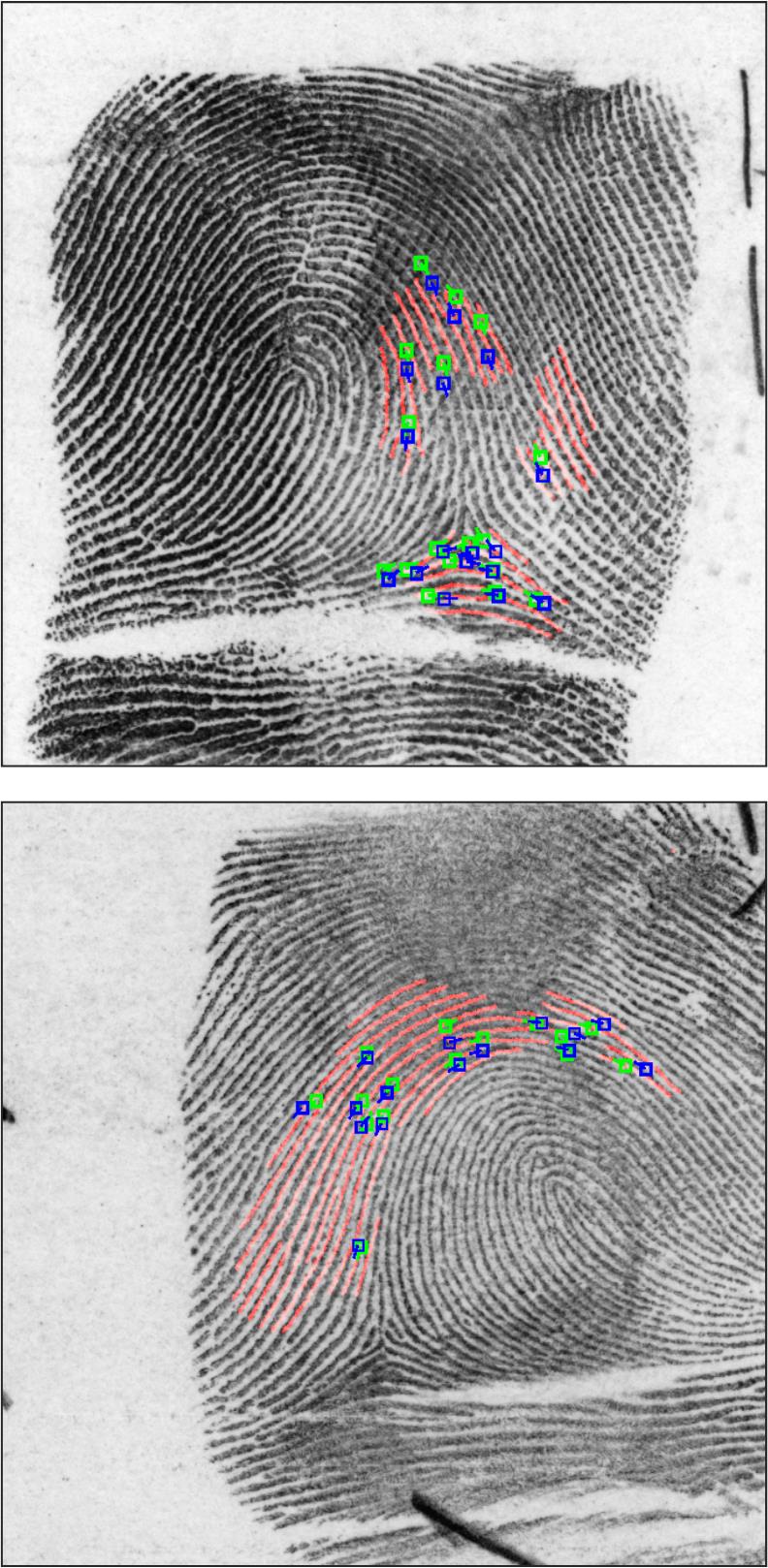}}
	 
\end{center}
	\caption{Precise registration helps improve registration accuracy. The first column shows the latent fingerprint to match with manually marked skeleton drawn on it. The second and third columns shows ground truth matching minutiae after coarse registration and precise registration, respectively, where minutiae on rolled fingerprints are marked in green, and registered minutiae on latent fingerprints are marked in blue.
 }
\label{fig: itermatching }
\end{figure}

Several examples are given to show the effect of precise registration in Fig. \ref{fig: itermatching }. In coarse registration, the intervals between sampling points are set very large to reduce the matching times of image patches in consideration of time efficiency. Therefore, the registered latent fingerprint is close to the correct result, but there is still a certain distance. To further reduce the registration error, the sampling points are denser  in the precise matching. Despite of more sampling points, the matching times are acceptable since each sampling point in latent fingerprint is only compared with 25 sampling points in rolled ones. The registration results  illustrates the effectiveness of precise registration.

\begin{table*}
\caption{Computational efficiency of fingerprint registration}
\begin{center}
\renewcommand\arraystretch{1.2}
\begin{tabular}{lcccccc}
\toprule
\multirow{2}{*}{Method} & \multirow{2}{*}{Stage} &\multicolumn{2}{c}{\textbf{Cao \textit{et al.}} \cite{cao2018end}} & \multirow{2}{*}{\textbf{Proposed-coarse}} & \multirow{2}{*}{\textbf{Proposed-precise}} & \multirow{2}{*}{\textbf{Dense registration} \cite{si2017dense}} \\
\cmidrule(lr){3-4} 
& & latent  &rolled & &\\
\midrule
Image preprocessing & Offline & 32.91 & 1.17 & 0.21 & 0 & -\\
\multirow{2}{*}{Descriptor extraction} & Offline & 137.48 (15.49+121.96+0.03)& 19.39 (10.23+9.13+0.03) & 22.73 & 0 & -\\
&Online &-&-&  0.93& 2.13 & -\\
Key point matching & Online & \multicolumn{2}{c}{0.001}& 0.14 & 0.29   & -\\
\midrule
 \multirow{2}{*}{Total time} & Offline &\multicolumn{2}{c}{ 190.95}& 23.87 & 0 &  -\\
 & Online &\multicolumn{2}{c}{0.001} & 1.07 & 2.42 & 3\\
\bottomrule
\end{tabular}
\end{center}
\label{tab:efficiency}
\end{table*}

\subsection{Computational Efficiency}

We compare the computational efficiency of proposed fingerprint registration method with Cao's method \cite{cao2018end} and dense registration \cite{si2017dense}. 
The speed of the dense registration  reported in their paper is cited for the comparison.
For the other two deep learning based approaches, the computational time taken for each method is related to the number of key points. Therefore, we report  the comparison results on NIST27 database. In the following, the comparison is made in both off-line and on-line stages.  

In the off-line stage, the image preprocessing step consists of ROI estimation, ridge orientation estimation, fingerprint enhancement, and minutiae extraction. Our method only requires ROI and ridge orientation estimation of rolled fingerprints, which is estimated by  FingerNet. 
The descriptor extraction step is applied for each key point patch. 
Cao \textit{et al.} extracts deep descriptors from each key point patch on latent and rolled fingerprints, respectively, followed by descriptor length reduction and product quantization  to improve the comparison speed, while we extract deep descriptors of very dense sampling points for on-line searching.

In the on-line stage, descriptor extraction step in our method adjusts the location and direction of sampling points with estimated transformation parameter and uses the descriptors of their nearest neighbor as their  deep descriptors. In the key point matching step, similarities between key points are computed, and global patch matching approach is applied in both methods to conduct the fingerprint comparison.
The time taken for each step is shown in Table \ref{tab:efficiency}. 
 
All the steps that require deep learning are conducted on NVIDIA GTX 1080 Ti. The key point matching in Cao's method is implemented with C++ while ours are implemented in MATLAB on a PC with 2.50 GHz CPU. The dense registration is implemented  in MATLAB and C.

As can be seen from Table \ref{tab:efficiency}, Cao's method takes a lot of time in the image preprocessing and extracting deep descriptors of latent fingerprints,  but the minutiae templates can be extracted offline. 
Compared with their method, ours  requires longer time online when registering a pair of fingerprints due to the two-stage registration. It takes about 3.49 seconds for one pair of fingerprints, which is comparable with the speed of dense registration. 
Therefore, the proposed method is more suitable for performing fine registration in the candidate lists generated by large database retrieval.

\section{Conclusion}

 In this paper, we propose a latent fingerprint registration method, which bypasses minutiae extraction and ridge orientation field estimation of latent fingerprints to avoid being affected by the error of feature extraction. Instead of minutiae, dense undirected sampling points are taken as the key points. All pairs of sampling points are compared to produce their similarities along with their relative spatial transformation. The sampling point correspondences are finally estimated by spectral clustering based global patch matching method.  To further reduce the registration error while not sacrificing efficiency,  an coarse-to-fine registration scheme is conducted.
The proposed latent fingerprint registration algorithm is tested on NIST27 database and MOLF database and compared with the state-of-the-art. Experimental results  show its better performance on latent fingerprints and stronger ability to handle fingerprints with poor quality.

The limitation of  the proposed approach is that (1) although the coarse-to-fine registration scheme helps increase the efficiency,  the local patch alignment and matching itself is time-consuming. (2)  It is inevitably affected by large skin distortion due to the performance constraints of local patch alignment and matching. (3) There is still room for improvement in the discrimination ability of the proposed deep key point descriptor.
Further direction includes increasing the computational efficiency while increasing the registration accuracy to better deal with local distortion,  and achieving end-to-end  fingerprint registration.

{\small
\bibliographystyle{unsrt}
\bibliography{Ref-new}
}

\end{document}